\documentclass{article} 
\usepackage{iclr2027_conference,times}

\usepackage{amsmath,amsfonts,bm}

\def\eqref#1{equation~\ref{#1}}

\def\1{\bm{1}}

\DeclareMathAlphabet{\mathsfit}{\encodingdefault}{\sfdefault}{m}{sl}
\SetMathAlphabet{\mathsfit}{bold}{\encodingdefault}{\sfdefault}{bx}{n}

\usepackage{hyperref}
\usepackage{url}
\usepackage{booktabs}
\usepackage{multirow}
\usepackage[table]{xcolor}
\usepackage{graphicx}
\usepackage{amssymb} 
\usepackage{booktabs, multirow, graphicx, amssymb}
\usepackage{pifont}
\usepackage{capt-of}
\usepackage{placeins}
\usepackage{float}  
\usepackage{makecell}
\usepackage{caption}

\title{FineSplat: Fine Patches, Local Groups, Detailed Splats}

\author{Hyebin Kim$^{\ast}$, Hwasik Jeong$^{\ast}$, Joonhyuk Park, Gyeongjin Kang, Jungwoo Kim, \\
\textbf{Younghyun Kim, Eunbyung Park$^{\dagger}$} \\ [0.5em]
Yonsei University}

\iclrfinalcopy 

\begin{document}
\maketitle
\lhead{Preprint}

\vspace{-2.5em}
{\centering \small
{\hypersetup{hidelinks}%
\href{https://hyebinny.github.io/FineSplat/}{\textcolor{blue}{\texttt{https://hyebinny.github.io/FineSplat/}}}}
\par}
\vspace{1em}

{\renewcommand{\thefootnote}{\fnsymbol{footnote}}
\footnotetext[1]{Equal contribution.}
\footnotetext[2]{Corresponding author.}}

\begin{figure}[h]
    \centering
    \vspace{-0.5cm}
    \includegraphics[width=\textwidth]{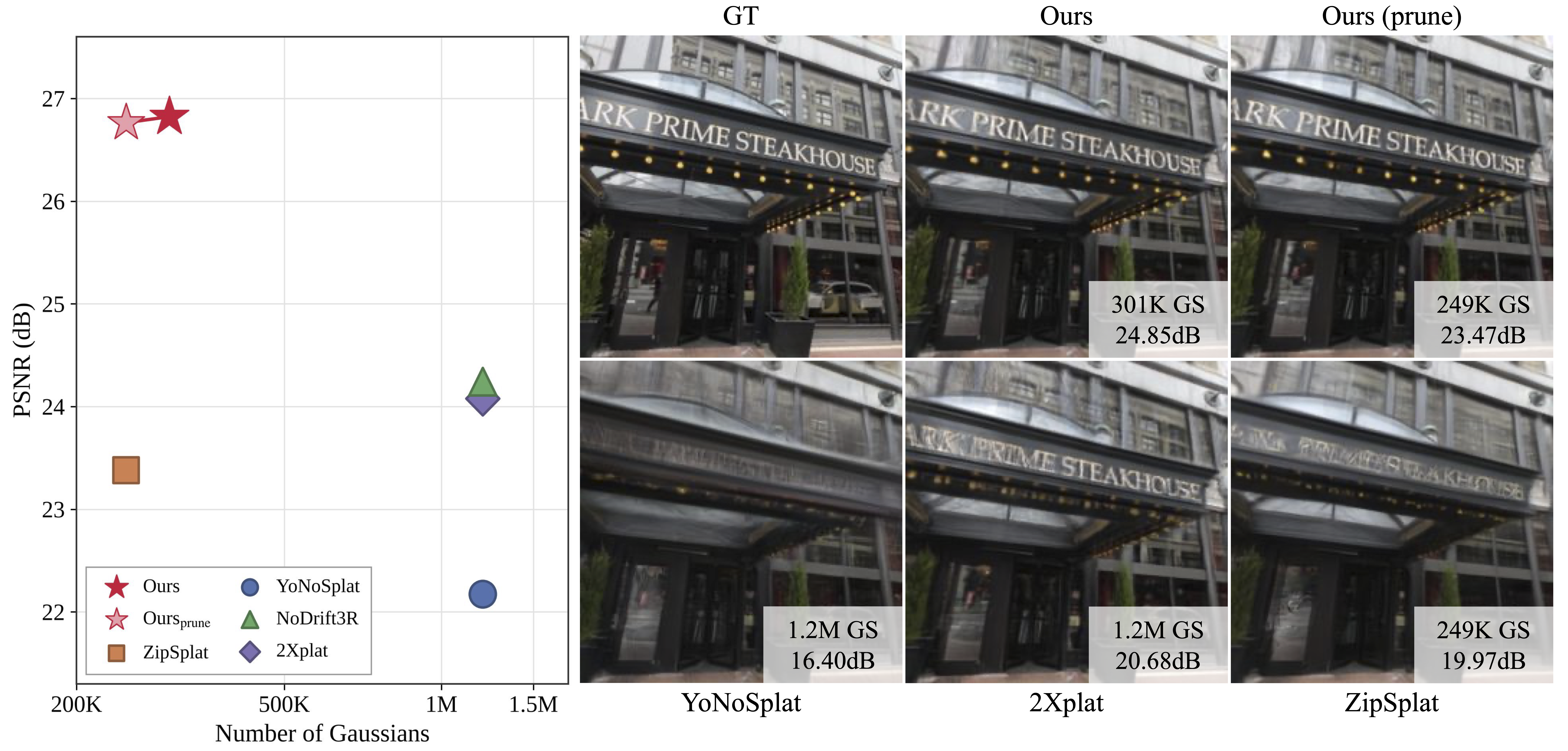}
    \caption{
    \textit{Left}: Number of Gaussians and average PSNR in the pose-free setting on DL3DV (224$\times$224 resolution, 24 context views).
    \textit{Right}: Qualitative comparison of novel view rendering in the pose-free setting with 24 context views.
    Our \textit{FineSplat} represents scenes more accurately and preserves the fine details while using the same or smaller Gaussian budgets.
    }
    \label{fig:teaser}
\end{figure}

\begin{abstract}
Feed-forward 3D Gaussian Splatting (3DGS) promises instant scene reconstruction with high-quality novel-view synthesis. Yet, fine-grained geometry and appearance remain difficult to recover, motivating us to revisit how feed-forward models represent, aggregate, and refine detailed visual information. We introduce \emph{FineSplat}, a feed-forward 3DGS framework built around fine-grained image patches. Since global multi-view attention becomes prohibitively expensive at such high token resolutions, we develop a \emph{dual-grouping} strategy that combines two complementary forms of locality: view-wise grouping enables efficient interaction among neighboring views, while 3D spatial grouping reconnects geometrically related tokens that may never have interacted under view-wise grouping by lifting them into the 3D scene space.
We further leverage reconstruction error as direct feedback, providing an explicit rendering-based signal to recover fine-grained structures and appearance, thereby overcoming the limitations of one-pass feed-forward prediction.
Finally, a compact Gaussian generation scheme preserves the benefits of fine-grained features without unnecessarily increasing the number of primitives. Together, these designs enable \emph{FineSplat} to recover richer geometry and appearance while remaining efficient and compact, achieving state-of-the-art results by a substantial margin, under the same or smaller Gaussian budgets than existing methods.

\end{abstract}
\section{Introduction}
3D Gaussian Splatting (3DGS)~\citep{kerbl20233dgs, yu2024mip, ye2025gsplat} has emerged as a powerful representation for high-quality and real-time novel view synthesis, combining an explicit 3D scene representation with efficient differentiable rasterization. Since conventional 3DGS relies on iterative per-scene optimization to achieve high reconstruction quality, it typically requires substantial computational resources for every new scene, limiting its scalability and applicability. To overcome these limitations, recent works have explored feed-forward 3DGS, where a single network forward pass directly predicts Gaussian representations from a set of input images. These approaches have rapidly advanced in both posed~\citep{charatan2024pixelsplat, chen2024mvsplat, ziwen2025longlrm, kang2026ilrm, kang2026mvp} and pose-free~\citep{zhang2025flare, ye2025noposplat, ye2026yonosplat, an2025c3g} settings, enabling fast reconstruction with promising generalization. However, replacing iterative per-scene refinement with a single feed-forward prediction places a much greater burden on the network, requiring it to resolve complex multi-view ambiguities and recover fine-grained scene details all at once. As a result, existing feed-forward approaches still exhibit a considerable reconstruction-quality gap compared with per-scene optimization methods, particularly in recovering fine-grained geometry and appearance details.
 
Many recent feed-forward 3DGS methods build upon powerful 3D foundation models, benefiting from their strong geometric priors and multi-view reasoning capabilities. However, these models commonly inherit the relatively coarse patch resolution of their geometry backbones, typically using $14\times14$ or $16\times16$ image patches~\citep{wang2025vggt, lin2026da3, wang2026pi3}. 
Such large patches provide broader spatial context that supports large-scale geometric reasoning and helps disambiguate cross-view correspondences, particularly when fine patches contain insufficiently distinctive local visual cues. However, compressing a broad image region into a single token can discard the high-frequency cues needed to reconstruct thin structures, sharp boundaries, textures, and other fine-grained details. Moreover, attention between coarse tokens operates at a limited spatial resolution, which can make it more difficult to precisely localize corresponding image regions across views. Finer patch tokens, particularly with attention between them, may therefore offer a promising avenue for preserving richer local visual cues and enabling more spatially precise cross-view interactions.
 
Yet, finer tokenization introduces a fundamental computational challenge. Most multi-view transformers rely on global self-attention across tokens from multiple input images, whose quadratic complexity quickly becomes prohibitive as the number of views, image resolution, and token density increase. A practical alternative is to restrict the attention scope. For example, MVP~\citep{kang2026mvp} partitions input views into small groups and performs fine-grained multi-view attention only within each group, substantially improving the efficiency of high-resolution multi-view reconstruction while maintaining strong accuracy. Such view-wise grouping is particularly effective when the inputs are sequentially captured and nearby views are likely to observe overlapping scene regions. However, the restriction is ultimately imposed in view space rather than scene space: tokens corresponding to the same 3D surface may fall into different view groups and therefore never directly interact. Consequently, any fixed view grouping inevitably creates blind spots, excluding potentially relevant correspondences simply because they lie across group boundaries.

In this work, we propose \emph{FineSplat}, which addresses this challenge through a \emph{dual-grouping} strategy that exploits two complementary forms of locality. First, we perform fine-grained multi-view interaction within small groups of neighboring views, enabling efficient attention over high-resolution tokens. We then lift the resulting tokens into 3D space using their predicted scene locations and regroup them according to 3D spatial proximity. This allows tokens from different, potentially distant view groups to interact when they correspond to nearby regions of the underlying 3D scene, recovering geometric relationships missed by view-wise grouping alone. By combining \emph{view-wise locality} with \emph{3D spatial locality}, \emph{FineSplat} enables rich interactions among fine-grained tokens without the prohibitive cost of global multi-view attention.

A faithful 3D reconstruction should reproduce the input images when rendered from their viewpoints, which provides a natural signal for assessing its quality.
Conventional per-scene optimization repeatedly exploits this signal, comparing rendered and observed images to identify residual errors and progressively correct the scene representation. A one-pass feed-forward model, in contrast, typically produces its reconstruction without an opportunity to inspect and correct its own errors. Motivated by this observation and recent work~\citep{xu2026resplat}, we explicitly leverage this readily available feedback as a direct signal in the second stage of our proposed framework. This provides the network with a second chance to recover structures and fine details that were missed in the first stage, without resorting to costly iterative optimization.
 
We further introduce a compact Gaussian generation scheme that predicts only a small number of Gaussians from each patch token, rather than densely generating pixel-aligned Gaussians. This design allows the model to retain the benefits of fine-grained image features without increasing the number of Gaussian primitives.
Together, the fine-grained representation, dual-grouping strategy, reconstruction error-driven correction, and compact Gaussian generation enable substantially improved reconstruction quality while maintaining an efficient and compact 3D representation.
Extensive experiments demonstrate that our method achieves state-of-the-art rendering quality by a substantial margin in both posed and pose-free settings, while using the same or fewer Gaussians than existing methods.
\section{Related Works}
\label{sec:related}
\paragraph{Feed-forward 3DGS and compact Gaussian representation.}
Feed-forward 3DGS methods amortize per-scene optimization~\citep{kerbl20233dgs} by directly predicting Gaussians from input images in a single forward pass.
One common design is pixel-aligned prediction, where each pixel is unprojected along its camera ray into one Gaussian, using epipolar attention~\citep{charatan2024pixelsplat}, cost volumes~\citep{chen2024mvsplat,xu2025depthsplat}, or large transformers~\citep{zhang2024gslrm,ziwen2025longlrm,kang2026mvp,wang2026tttlrm}.
This keeps the geometry well-conditioned and helps preserve fine appearance, but the number of Gaussians grows linearly with resolution and view count, often producing many redundant primitives in overlapping regions~\citep{park2025ecosplat,kim2026f4splat,RoSplat}.
In contrast, non-pixel-aligned methods predict a compact set of Gaussians from learnable tokens, clustered scene features, 3D anchors, or voxels~\citep{ren2026tokengs,an2025c3g,veicht2026zipsplat, wang2026volsplat,moreau2026offthegrid,zhang2026anchorsplat,itkin2026globalsplat,zhang2026sparsesplat,wang2026zpressor}, but lose the explicit pixel-ray correspondence, which tends to hurt fine detail and makes optimization harder.
Between these extremes, patch-aligned methods predict Gaussians at a reduced spatial resolution~\citep{kang2026ilrm,xu2026resplat}.
In our framework, we extend this patch-aligned approach by generating only a small number of Gaussians along the rays of each sub-patch, and further compact the representation by pruning Gaussians according to their rendering contribution to the context views.

\paragraph{Pose-free feed-forward 3DGS and geometry foundation models.}
Geometry foundation models~\citep{wang2024dust3r,leroy2024mast3r,wang2025vggt,wang2026pi3,lin2026da3} estimate camera poses and dense geometry directly from unposed images.
Building on such models, pose-free methods jointly predict camera poses and Gaussians~\citep{ye2025noposplat,zhang2025flare,jiang2025anysplat,ye2026yonosplat,kang2025selfsplat}, mostly in a pixel-aligned manner on top of the geometry backbone.
However, these backbones operate on relatively coarse image tokens, which are well suited for global geometric reasoning but may discard local appearance cues important for fine-grained Gaussian reconstruction.
To compensate, existing methods often fine-tune the backbone end-to-end so that its pretrained features adapt to appearance reconstruction.
This adaptation is computationally expensive, and the model still relies on representations shaped by the coarse tokenization, which may limit its ability to recover fine-grained appearance.
Recent works instead decouple geometry estimation from appearance modeling~\citep{zhang2025flare,jeong2026twoxplat}.
We follow this decoupled design for pose-free reconstruction, using a 3D foundation model, such as DA3~\citep{lin2026da3}, to predict the camera poses while \emph{FineSplat} performs fine-grained Gaussian prediction.

\paragraph{Combining image-space and 3D locality.}
Many useful cues for 3D reconstruction, such as dense image features and rendering errors, are naturally defined in image space, where they are grouped by proximity in the image plane rather than in the underlying scene.
Several works therefore bring such image-space cues into 3D and process them according to 3D spatial locality.
ODIN~\citep{jain2024odin} alternates 2D within-view layers with 3D cross-view attention over unprojected features, and DITR~\citep{knaebel2026ditr} projects DINOv2~\citep{oquab2023dinov2} features onto 3D points and injects them into a Point Transformer V3~\citep{wu2024ptv3} backbone.
MoGe-3~\citep{kong2026moge} lifts monocular geometry into a sparse voxel volume guided by image features to recover sharp details near depth boundaries.
In feed-forward 3DGS, ATSplat~\citep{cho2026atsplat} unprojects patch features into sparse 3D anchors that gather context from their 3D neighbors, and expands them where rendering errors are predicted to be high.
Our second stage lifts fine tokens from all views into 3D and applies serialized attention over their 3D neighborhoods, so that tokens separated by view grouping can interact.

\paragraph{Iterative refinement.}
Iterative minimization of an objective has long been a standard approach to solving vision problems~\citep{lucas1981iterative,sun2010secrets}, and learning-based methods have sought to emulate this process by repeatedly updating an initial estimate.
Such learned iterative refinement often outperforms single-step regression and has been successfully applied to a broad range of applications, such as optical flow~\citep{teed2020raft}, human pose estimation~\citep{newell2016hourglass,carreira2016human}, stereo matching~\citep{lipson2021raft}, SLAM~\citep{teed2021droid}, structure-from-motion~\citep{li2025megasam}, multi-view stereo~\citep{wang2022itermvs}, and multi-view 3D reconstruction~\citep{dejaview2026}.
For Gaussian splatting, prior works refine Gaussians with a single-step or multiple-step network~\citep{chen2025splatformer, nam2025generative, kang2026ilrm}, learn updates from rendering gradients~\citep{chen2024g3r,liu2025quicksplat, liu2026diff3r, pearl2026learn2splat} or rendering errors~\citep{xu2026resplat}, or fine-tune Gaussian tokens for each scene~\citep{ren2026tokengs}.

\begin{figure}[t]
    \centering
    \includegraphics[width=\textwidth]{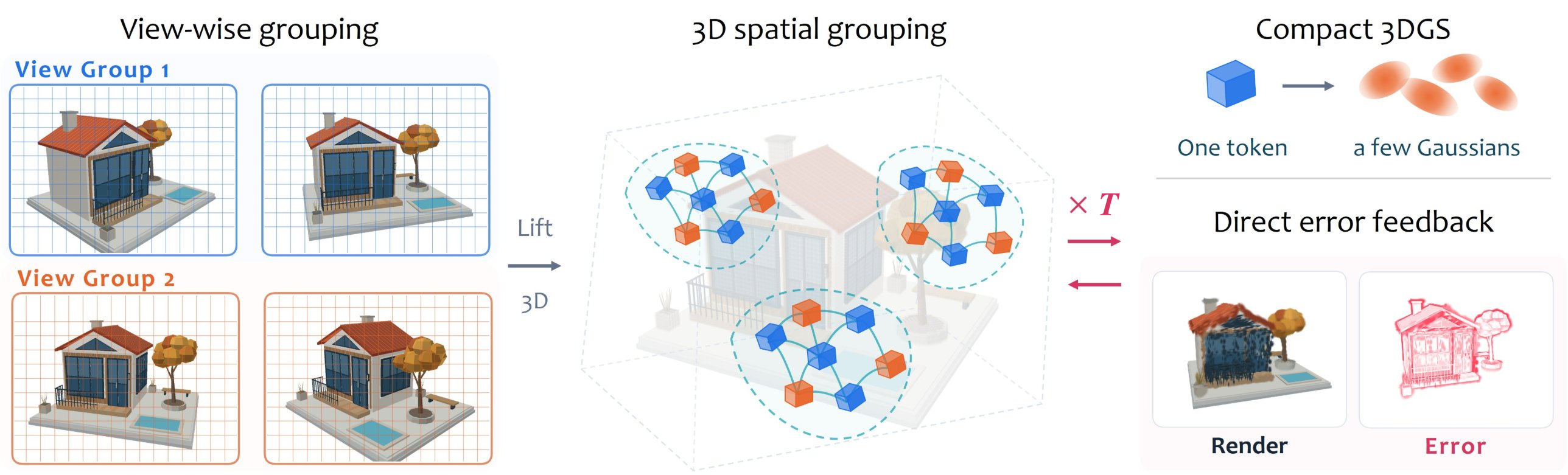}
    \caption{
    \textbf{Method Overview.} \emph{FineSplat} first performs view-wise grouping and applies self-attention within each group, enabling efficient processing of fine-grained tokens. 3D spatial grouping lifts tokens into 3D space and enables interactions among spatially neighboring tokens, including those from different view groups, to enhance cross-view information exchange. Compact Gaussian generation predicts fewer Gaussians per token than existing pixel-aligned methods, maintaining a compact 3DGS representation. Finally, direct error feedback incorporates reconstruction errors obtained by rendering the input views to further refine token representations.
    }
    \label{fig:main_figure}
\end{figure}

\section{Method}
Given $N$ context views with or without known camera poses, \emph{FineSplat} predicts a set of 3D Gaussians in a feed-forward manner.
Stage~1 divides each context view into fine-grained $4\times4$ patches and encodes them with attention of varying scope, where group-wise attention over a few views keeps fine-grained multi-view reasoning affordable.
Stage~2 decodes these tokens into an initial set of Gaussians, renders them
back into the context views, and uses the resulting error maps to update the tokens through 3D spatial grouping, which lets tokens that fall in different view groups interact once they are lifted to 3D.
This dual-grouping strategy makes fine-grained encoding tractable, with 3D spatial grouping complementing view-wise grouping by connecting tokens across view groups.  Figure~\ref{fig:main_figure} gives an overview of the two stages.

\subsection{Stage~1: View-wise token grouping}
Global multi-view attention becomes prohibitively expensive with fine-grained tokens. In the first stage, we restrict the attention scope and adopt a view-wise token grouping strategy. Specifically, we leverage the multi-view pyramid architecture (MVP)~\citep{kang2026mvp}.
It performs fine-grained attention within small groups of views, while allowing information exchange across all views through coarser global tokens. This balances local detail preservation with global multi-view reasoning.
It organizes its encoder as a three-level pyramid, and we denote the feature map of level $\ell\in\{1,2,3\}$ as $F^{\ell}$, from the finest ($F^{1}$) to the coarsest ($F^{3}$).
In more detail, given $N$ input images ${I}\in\mathbb{R}^{N\times H\times W\times 3}$ and patch size $p$, the first stage fuses the three levels of coarse-to-fine features into $M=\frac{NHW}{p^2}$ fine-grained tokens ${z}^{(0)}\in\mathbb{R}^{M \times d}$,
\begin{equation}
    {z}^{(0)} = \mathrm{fuse}\big(\mathrm{up}(\mathrm{fuse}(\mathrm{up}(F^{3}) + F^{2})) + F^{1}\big),
    \label{eq:stage1_tokens}
\end{equation}
where $\mathrm{up}(\cdot)$ denotes bilinear upsampling and $\mathrm{fuse}(\cdot)$ a convolutional fusion block, which are used to match the token resolutions.

The fixed view grouping strategy inevitably leaves blind spots: fine-grained tokens in different groups cannot directly interact, potentially missing relevant observations from other input views. Thus, view-wise grouping alone cannot fully realize fine-grained multi-view reasoning over all input views. To recover these missing interactions, we pass the fine-grained tokens to the subsequent 3D spatial grouping stage.

\subsection{Stage~2: 3D spatial token grouping}

Starting from the fine-grained tokens ${z}^{(0)}$ produced by the first stage, we first decode an initial set of 3D Gaussians. Specifically, a lightweight Gaussian head $\mathcal{D}$, implemented as a linear decoder, maps each token to $K$ Gaussian primitives and their attributes. We then use the predicted Gaussian centers to lift the corresponding tokens into 3D space, where tokens are regrouped according to their spatial proximity in 3D world coordinates.
We then perform self-attention within each 3D group, enabling fine-grained interactions between geometrically related tokens even when they originate from different view groups. To further refine the current estimate, we augment this interaction with reconstruction-error information obtained by rendering the input views. The updated tokens are subsequently decoded into a new set of Gaussian primitives, and this process of Gaussian prediction, 3D grouping, error-guided interaction, and token update is repeated for multiple steps.

\paragraph{Lifting tokens to 3D.}
To reason about the tokens in a common 3D space, we anchor each token at the mean position of the Gaussians it produces. Given the $j$-th token at the $i$-th step, $z_j^{(i)} \in \mathbb{R}^{d}$,
\begin{equation}
    \{{\mu}_{jk}^{(i)}\}_{k=1}^{K} = \mathcal{D}(z_j^{(i)}), \quad {p}_j^{(i)} = \frac{1}{K}\sum_{k=1}^{K}{\mu}_{jk}^{(i)},
    \label{eq:token_lifting}
\end{equation}
where ${\mu}_{jk}^{(i)}\in\mathbb{R}^3$ is the position of each predicted Gaussian. Each token is then projected by a linear layer, 
yielding a point cloud $\{({p}_j^{(i)}, {z}_j^{(i)})\}_{j=1}^{M}$ whose points carry the fine-grained tokens as features.

\paragraph{Error feature maps.}
To measure how well the current Gaussians explain the observations, we render them into every input view. For the $n$-th context view, we define a signed error map between the input image ${I}_n$ and the rendered image $\hat{I}_n$, computed as ${E}_n =  \hat{I}_n - {I}_n$.
To complement this pixel-level discrepancy with structural cues, we additionally compute a feature-level difference $\Delta{\phi}_n = \phi(\hat{I}_n) - \phi({I}_n)$, where $\phi(\cdot)$ is a pretrained ResNet-18~\citep{he2016deep} encoder. These two terms are combined with the finest-scale features from the first stage, ${F}^1_n$, which provide the image context in which the errors occur. We obtain the error feature map as follows,
\begin{equation}
    {F}_n^{\mathrm{err}} = \big[\,\mathcal{E}_{\mathrm{err}}({E}_n, \Delta{\phi}_n)\,;\; \mathcal{E}_{\mathrm{feat}}({F}^{1}_n)\,\big] \in \mathbb{R}^{\frac{H}{p}\times\frac{W}{p} \times d_e},
    \label{eq:error_feature_map}
\end{equation}
where ${F}^1_n$ denotes the finest-scale features from the first stage for an input view $n$, $\mathcal{E}_{\mathrm{err}}$ and $\mathcal{E}_{\mathrm{feat}}$ are linear embedding layers, and $[\cdot;\cdot]$ represents the feature concatenation.

\paragraph{Gathering errors across views.}
A Gaussian affects the rendering of every view in which it is visible, not only the view from which it was predicted. We therefore gather error features for each token from both its own view and the other views that observe it.
Let $\pi_n(\cdot)$ denote the projection onto the error feature map of view $n$, and let $v_{jn}\in\{0,1\}$ indicate whether ${p}_j^{(i)}$ lies inside the view $n$'s frustum and passes an occlusion test against the rendered depth.
The aggregated error feature is obtained by averaging the bilinearly sampled features over the visible views,
\begin{equation}
    {e}_j^{\mathrm{agg}} = \frac{\sum_{n=1}^{N} v_{jn}\,{F}_n^{\mathrm{err}}\big(\pi_n({p}_j^{(i)})\big)}{\sum_{n=1}^{N} v_{jn}},
    \label{eq:error_agg}
\end{equation}
where ${F}_n^{\mathrm{err}}(\cdot)$ reads a feature given a projected coordinate. In addition, we take the error feature at the patch from which ${z}_j^{(i)}$ originates, denoted as ${e}_j^{\mathrm{self}}$. 
While ${e}_j^{\mathrm{self}}$ provides a direct signal from the source view, ${e}_j^{\mathrm{agg}}$ summarizes how consistently the token explains the other views.
The two are fused into the final error feature ${e}_j = \mathrm{Linear}\big([{e}_j^{\mathrm{self}}; {e}_j^{\mathrm{agg}}]\big)$.

\paragraph{3D spatial token grouping and update.}
Each lifted point is assigned the feature ${h}_j = \mathrm{Linear}\big([{z}_j^{(i)}; {e}_j]\big)$.
Neighboring tokens in 3D often originate from different views and describe the same surface, so correcting them jointly is more effective than correcting each token independently. We thus exchange information among nearby points using serialized attention~\citep{wu2024ptv3}, which orders the points along a space-filling curve, partitions them into groups of $S$ consecutive points, and applies self-attention within each group.
Compared with explicit $k$-nearest-neighbor search, this provides local 3D context at a much lower cost and scales favorably with the number of tokens.
The updated features are projected to a token residual by a linear layer,
\begin{equation}
    \Delta z_j^{(i)} = \mathrm{Linear}\big(\mathrm{SerialAttn}({h_j})\big), \qquad
    z_j^{(i+1)} = z_j^{(i)} + \Delta z_j^{(i)}.
    \label{eq:token_update}
\end{equation}
Then, the updated tokens are decoded into Gaussians with the shared Gaussian head $\mathcal{D}$.
Our Stage~2 module is applied for $T=4$ steps, starting from ${z}^{(0)}$: each step takes the tokens from the previous one, re-renders the context views, recomputes the error features, and regroups the tokens at their updated 3D positions.

\subsection{Patch-aligned Sub-pixel Gaussian Reconstruction}

To obtain a dense Gaussian representation while preserving the spatial structure of image features, we predict four Gaussians for each $4\times4$ fine-grained token, assigning each Gaussian to one of four non-overlapping $2\times2$ sub-patches.
Rather than constraining a Gaussian to the fixed center ray of its assigned sub-patch, we learn a two-dimensional image-space offset $\Delta = (\Delta_x, \Delta_y)$ that adaptively shifts its ray within the corresponding region. Specifically, for a sub-patch with stride $s=2$, we bound the offset as $\delta_x=\frac{s}{2}\tanh{(\Delta_x)}, \delta_y=\frac{s}{2}\tanh{(\Delta_y)}$, such that $\delta_x, \delta_y \in [-s/2, s/2]$. This guarantees that each Gaussian remains within its designated $2\times2$ sub-patch while allowing continuous sub-pixel localization.
Given the center $(u_c, v_c)$ of each sub-patch, the displaced image coordinate is $(u,v)=(u_c,v_c)+\delta$, which is unprojected using the predicted camera intrinsics.
We then transform and normalize this direction using the camera-to-world rotation, with the camera translation serving as the corresponding ray origin.
\section{Experiments}
\label{sec:experiments}
\subsection{Implementation Details}
\paragraph{Setup.}

We train a single model only on the DL3DV~\citep{ling2024dl3dv} dataset, sampling the number of input context views from $\{6, 12, 24, 32\}$ for each training scene, and use this one checkpoint for all evaluations \textbf{\textit{without dataset-specific fine-tuning}}.
On DL3DV, we evaluate on 140 test scenes with $N \in \{6, 12, 24\}$ context views, selected via farthest point sampling with maximum temporal frame gaps of 50, 100, and 150, respectively.
We further evaluate zero-shot generalization on 1{,}600 RealEstate10K (RE10K)~\citep{zhou2018stereo} test scenes with $N = 6$, and on Mip-NeRF~360~\citep{barron2022mip} with $N \in \{32, 64, 128\}$, following the view indices from~\citet{veicht2026zipsplat}.
Mip-NeRF~360 is a demanding target for a model trained on DL3DV: it covers both indoor and outdoor scenes, its scenes are unbounded, its cameras orbit the scene rather than sweep through it, and it uses far more context views than the 32 seen during training.
We hold out 8 target views on all three datasets.
For a fair comparison, each method encodes and renders at its native training resolution, and all rendered outputs are resized to $224 \times 224$ for evaluation, following~\citet{ye2026yonosplat} (252 $\times$ 252 results are provided in Appendix~\ref{sec:eval_reslution}).
We report PSNR, SSIM~\citep{wang2004image}, and LPIPS~\citep{zhang2018unreasonable}, averaged over all test scenes.

\paragraph{Supervision and scene normalization.}
\emph{FineSplat} is \textbf{\textit{trained without any depth supervision}}, using only a mean squared error term and a perceptual term on the rendered views.
Several of the baselines instead rely on depth: AnySplat~\citep{jiang2025anysplat}, DepthSplat~\citep{xu2025depthsplat}, NoDrift3R~\citep{sun2026nodrift3r}, DA3~\citep{lin2026da3}, and ZipSplat~\citep{veicht2026zipsplat} all use a depth objective during training.
Since feed-forward reconstruction is scale-ambiguous, we bring each scene to a canonical scale using only the context camera poses: poses are expressed in the mean context camera frame and translations are rescaled so that all context camera centers lie within $[-1,1]^3$.
The same transform is applied to the target poses, both ground-truth and DA3-predicted.
ZipSplat instead derives the scene scale from depth at evaluation time, unprojecting per-view depth maps predicted by DA3 and taking the median distance of the resulting points to the first context camera.
Details of the scene scale normalization are provided in Appendix~\ref{sec:scene_scale}.
Since we evaluate on the same Mip-NeRF~360 scenes and view indices as ZipSplat, the two differ only in how this scale is obtained: its normalization depends on a dense depth prior, whereas ours needs only camera positions and transfers directly to predicted poses.

\paragraph{Baselines.} 
We compare against the pose-free methods AnySplat~\citep{jiang2025anysplat}, C3G~\citep{an2025c3g}, 
DA3~\citep{lin2026da3}, YoNoSplat~\citep{ye2026yonosplat}, NoDrift3R~\citep{sun2026nodrift3r}, 2Xplat~\citep{jeong2026twoxplat}, and ZipSplat~\citep{veicht2026zipsplat}, as well as NoPoSplat~\citep{ye2025noposplat}, which additionally requires ground-truth camera intrinsics. For pose-dependent methods, we compare against DepthSplat~\citep{xu2025depthsplat} and MVP~\citep{kang2026mvp}.

\subsection{Novel view synthesis}
\paragraph{Benchmark performance.}
\begin{figure}[t]
    \centering
    \begin{minipage}{\textwidth}
        \begin{center}
        \captionof{table}{
            Novel view synthesis results on DL3DV with 6, 12, and 24 input views.
        }
        \label{tab:dl3dv_main}
\scriptsize
\setlength{\tabcolsep}{3pt}
\resizebox{\textwidth}{!}{%
\begin{tabular}{lcc|cccc|cccc|cccc}

\toprule

    \multirow{2}{*}{Method}
    & \multirow{2}{*}{$p$}
    & \multirow{2}{*}{$k$}
    & \multicolumn{4}{c|}{6 views}
    & \multicolumn{4}{c|}{12 views}
    & \multicolumn{4}{c}{24 views} \\
    \cmidrule(lr){4-7}
    \cmidrule(lr){8-11}
    \cmidrule(lr){12-15}
    &
    &
    &
    \# GS & PSNR $\uparrow$ & SSIM $\uparrow$ & LPIPS $\downarrow$
    & \# GS & PSNR $\uparrow$ & SSIM $\uparrow$ & LPIPS $\downarrow$
    & \# GS & PSNR $\uparrow$ & SSIM $\uparrow$ & LPIPS $\downarrow$ \\

\midrule


    DepthSplat
    & \checkmark & \checkmark
    & 393K & 23.418 & 0.797 & 0.136
    & 786K & 21.911 & 0.753 & 0.179
    & 1.6M & 20.088 & 0.690 & 0.240 \\

    YoNoSplat
    & \checkmark & \checkmark
    & 301K & 24.717 & 0.817 & 0.139
    & 602K & 23.285 & 0.773 & 0.177
    & 1.2M & 22.664 & 0.758 & 0.192 \\

    MVP
    & \checkmark & \checkmark
    & 301K & 24.314 & 0.798 & 0.181
    & 602K & 23.432 & 0.775 & 0.210
    & 1.2M & 23.106 & 0.768 & 0.221 \\

    2Xplat
    & \checkmark & \checkmark
    & 301K & 25.217 & 0.827 & 0.150
    & 602K & 24.592 & 0.813 & 0.169
    & 1.2M & 24.354 & 0.811 & 0.174 \\

    DA3
    & \checkmark & \checkmark
    & 1.5M & 19.955 & 0.591 & 0.212
    & 3.0M & 20.263 & 0.602 & 0.213
    & 6.1M & 20.527 & 0.626 & 0.212 \\


    ZipSplat
    & \checkmark & \checkmark
    & 62K & 23.943 & 0.775 & 0.169
    & 124K & 22.730 & 0.718 & 0.198
    & 249K & 22.536 & 0.715 & 0.201 \\





    Ours
    & \checkmark & \checkmark
    & 75K & \textbf{27.325} & \textbf{0.871} & \textbf{0.115}
    & 151K & \textbf{27.140} & \textbf{0.868} & \textbf{0.121}
    & 301K & \textbf{27.233} & \textbf{0.873} & \textbf{0.119} \\

    Ours$_\mathrm{prune}$
    & \checkmark & \checkmark
    & 62K & 27.165 & 0.866 & 0.126
    & 124K & 27.031 & 0.864 & 0.129
    & 249K & 27.164 & 0.871 & 0.125 \\
    




\midrule

    NoPoSplat
    & & \checkmark
    & 393K & 22.766 & 0.743 & 0.179
    & 786K & 19.380 & 0.563 & 0.318
    & 1.6M & 17.860 & 0.495 & 0.397 \\

    YoNoSplat
    & & \checkmark
    & 301K & 24.887 & 0.819 & 0.138
    & 602K & 23.149 & 0.758 & 0.183
    & 1.2M & 22.354 & 0.731 & 0.205 \\

    2Xplat
    & & \checkmark
    & 301K & 25.210 & 0.825 & 0.152
    & 602K & 24.337 & 0.798 & 0.177
    & 1.2M & 24.049 & 0.792 & 0.188 \\








\midrule

    AnySplat
    & &
    & 951K & 19.027 & 0.554 & 0.235
    & 1.8M & 18.940 & 0.549 & 0.262
    & 3.2M & 19.703 & 0.596 & 0.249 \\


    YoNoSplat
    & &
    & 301K & 24.531 & 0.804 & 0.142
    & 602K & 22.933 & 0.746 & 0.187
    & 1.2M & 22.174 & 0.720 & 0.209 \\

    NoDrift3R
    & &
    & 301K & 24.922 & 0.826 & 0.127
    & 602K & 24.250 & 0.797 & 0.141
    & 1.2M & 24.242 & 0.794 & 0.142 \\


    2Xplat
    & &
    & 301K & 25.227 & 0.825 & 0.151
    & 602K & 24.360 & 0.799 & 0.177
    & 1.2M & 24.077 & 0.793 & 0.185 \\

    C3G
    & &
    & 2K & 18.608 & 0.481 & 0.400
    & 2K & 16.138 & 0.392 & 0.548
    & 2K & 15.055 & 0.358 & 0.601 \\

    DA3
    & &
    & 1.5M & 23.065 & 0.791 & 0.152
    & 3.0M & 22.742 & 0.773 & 0.166
    & 6.1M & 22.649 & 0.767 & 0.173 \\

    ZipSplat
    & &
    & 62K & 24.408 & 0.789 & 0.166
    & 124K & 23.497 & 0.749 & 0.192
    & 249K & 23.380 & 0.749 & 0.193 \\




    Ours
    &  & 
    & 75K & \textbf{26.997} & \textbf{0.861} & \textbf{0.118}
    & 151K & \textbf{26.778} & \textbf{0.857} & \textbf{0.125}
    & 301K & \textbf{26.822} & \textbf{0.857} & \textbf{0.127} \\        

    Ours$_\mathrm{prune}$
    &  & 
    & 62K & 26.860 & 0.857 & 0.129
    & 124K & 26.686 & 0.853 & 0.134
    & 249K & 26.763 & 0.855 & 0.133 \\



\bottomrule

\end{tabular}%
}

        \end{center}
    \end{minipage}

    \vspace{0.5em}

    \begin{minipage}{\textwidth}
        \begin{center}
        \includegraphics[width=\textwidth]{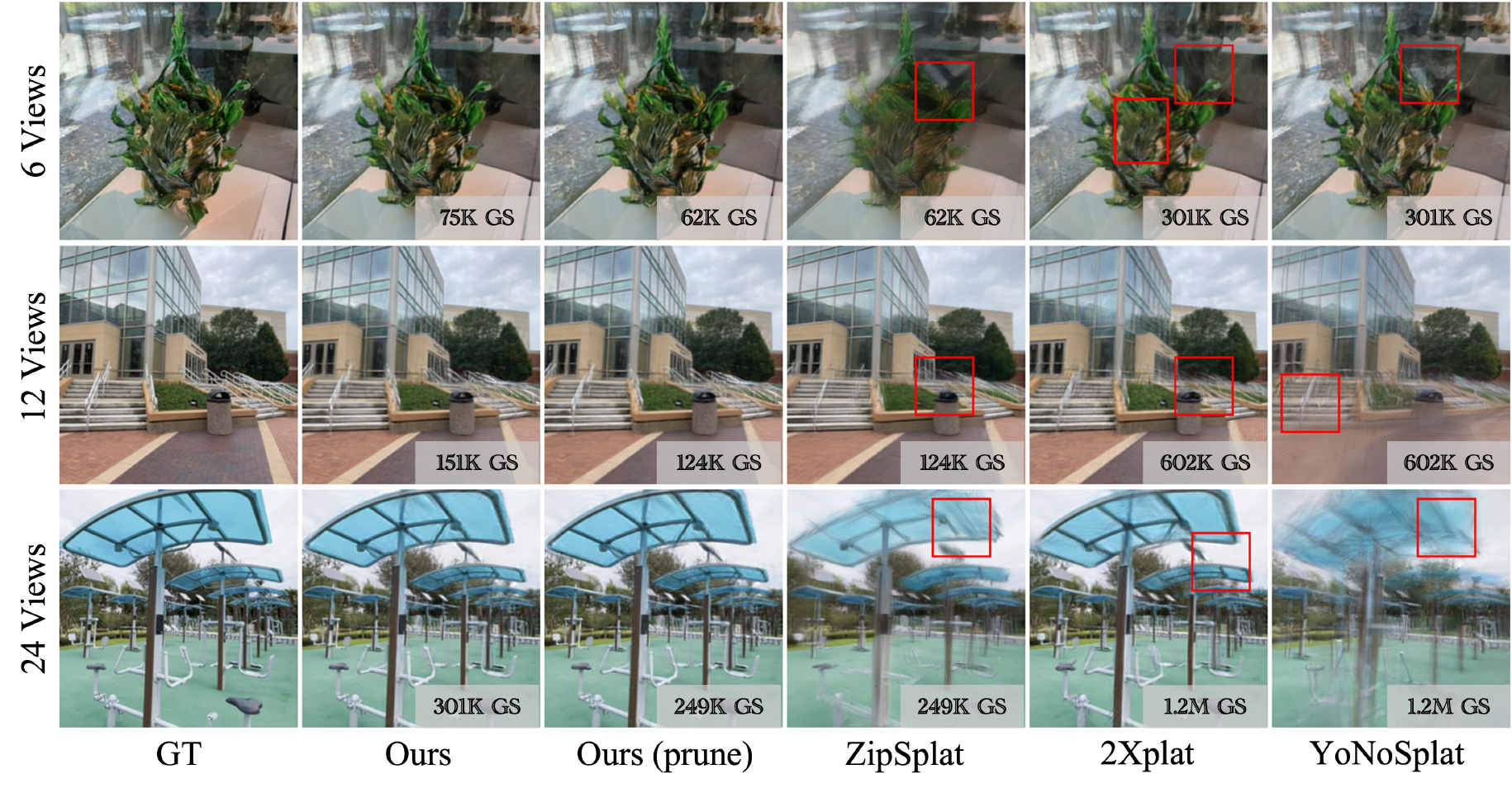}
        \captionof{figure}{
            Qualitative comparison on DL3DV with varying numbers of input views.
        }
        \label{fig:dl3dv_main_qualitative}
        \end{center}
    \end{minipage}
\end{figure}

As shown in Table~\ref{tab:dl3dv_main}, \emph{FineSplat} consistently outperforms previous state-of-the-art methods in both posed and fully pose-free settings across all input-view configurations. Compared with 2Xplat, ours improves PSNR by up to $+2.88$ dB in the posed setting and $+2.75$ dB in the pose-free setting, while using about 4 times fewer Gaussians. Note that we compare with the 2Xplat variant that does not leverage view-dependent opacity for a fair comparison. In addition, ours remains substantially more stable as the number of views increases, which we attribute to our two-stage dual-grouping formulation with fine-grained tokens that better disambiguates redundant and highly overlapping regions across views, reducing conflicting Gaussian predictions.
Moreover, some methods achieve competitive PSNR but relatively high LPIPS, whereas ours achieves both high PSNR and low LPIPS. We attribute this to predicting Gaussians along the rays of fine-grained patches, which better preserves the fine structure of the scene beyond photometric accuracy.
Figure ~\ref{fig:dl3dv_main_qualitative} shows representative qualitative results.

\paragraph{Cross-dataset generalization.}
\begin{figure}[t]
    \centering
    \begin{minipage}[t]{0.58\textwidth}
        \vspace{0pt}
        \centering
        \captionof{table}{
            Cross-dataset generalization results on RE10K with 6 input views.
            Methods shown in \textcolor{gray}{gray} are trained on RE10K.
        }
        \label{tab:re10k_main}
        \begin{center}       
\scriptsize
\setlength{\tabcolsep}{3pt}
\resizebox{\linewidth}{!}{%
\begin{tabular}{lcc|cccc}

\toprule

    \multirow{2}{*}{Method}
    & \multirow{2}{*}{$p$}
    & \multirow{2}{*}{$k$}
    & \multicolumn{4}{c}{6 views} \\
    \cmidrule(lr){4-7}
    & & &
    \# GS & PSNR $\uparrow$ & SSIM $\uparrow$ & LPIPS $\downarrow$ \\

\midrule

    \textcolor{gray}{DepthSplat}
    & \textcolor{gray}{\checkmark} & \textcolor{gray}{\checkmark}
    & \textcolor{gray}{393K} & \textcolor{gray}{24.156} & \textcolor{gray}{0.846} & \textcolor{gray}{0.145} \\

    \textcolor{gray}{YoNoSplat}
    & \textcolor{gray}{\checkmark} & \textcolor{gray}{\checkmark}
    & \textcolor{gray}{301K} & \textcolor{gray}{25.037} & \textcolor{gray}{0.848} & \textcolor{gray}{0.134} \\

    \textcolor{gray}{2Xplat}
    & \textcolor{gray}{\checkmark} & \textcolor{gray}{\checkmark}
    & \textcolor{gray}{301K} & \textcolor{gray}{26.374} & \textcolor{gray}{0.875} & \textcolor{gray}{0.136} \\

    \textcolor{gray}{ZipSplat}
    & \textcolor{gray}{\checkmark} & \textcolor{gray}{\checkmark}
    & \textcolor{gray}{62K} & \textcolor{gray}{24.890} & \textcolor{gray}{0.815} & \textcolor{gray}{0.152} \\



    DA3
    & \checkmark & \checkmark
    & 1.5M & 19.693 & 0.645 & 0.237 \\

    


    Ours
    & \checkmark & \checkmark
    & 75K & \textbf{27.394} & \textbf{0.886} & \textbf{0.116} \\
    
    Ours$_\mathrm{prune}$
    & \checkmark & \checkmark
    & 62K & 27.330 & 0.885 & 0.121 \\





\midrule


    \textcolor{gray}{YoNoSplat}
    & &
    & \textcolor{gray}{301K} & \textcolor{gray}{24.571} & \textcolor{gray}{0.823} & \textcolor{gray}{0.144} \\

    \textcolor{gray}{NoDrift3R}
    & &
    & \textcolor{gray}{301K} & \textcolor{gray}{25.736} & \textcolor{gray}{0.876} & \textcolor{gray}{0.128} \\


    \textcolor{gray}{2Xplat}
    & &
    & \textcolor{gray}{301K} & \textcolor{gray}{27.239} & \textcolor{gray}{0.881} & \textcolor{gray}{0.126} \\

    \textcolor{gray}{C3G}
    & &
    & \textcolor{gray}{2K} & \textcolor{gray}{20.215} & \textcolor{gray}{0.638} & \textcolor{gray}{0.316} \\

    \textcolor{gray}{ZipSplat}
    & &
    & \textcolor{gray}{62K} & \textcolor{gray}{25.474} & \textcolor{gray}{0.825} & \textcolor{gray}{0.155} \\


    DA3
    & &
    & 1.5M & 20.631 & 0.712 & 0.228 \\




    
    Ours
    & &
    & 75K & \textbf{26.554} & \textbf{0.854} & \textbf{0.127} \\

    Ours$_\mathrm{prune}$
    & &
    & 62K & 26.514 & 0.853 & 0.132 \\





\bottomrule

\end{tabular}%
}

        \end{center}
    \end{minipage}%
    \hfill    
        \begin{minipage}[t]{0.41\textwidth}
        \vspace{0pt}
        \centering
        \captionsetup{skip=2pt}
        \includegraphics[width=\linewidth]{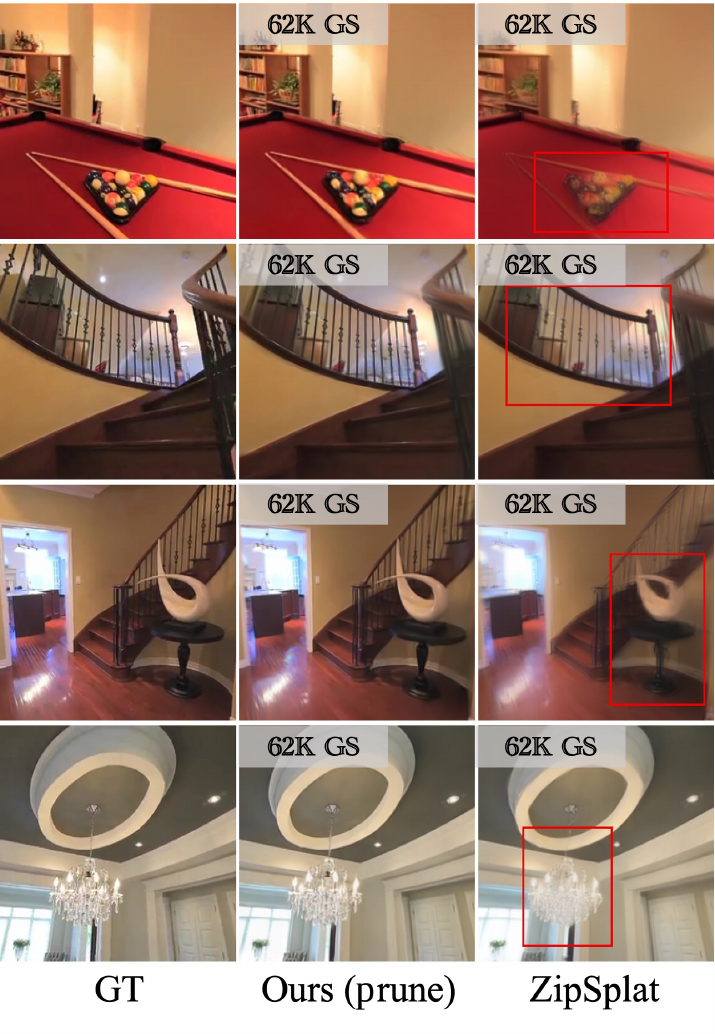}
        \captionof{figure}{
            Qualitative comparison on RE10K with 6 input views.
        }
        \label{fig:re10k_main_qualitative}
        \end{minipage}
\end{figure}

\begin{figure}[t]
    \centering
    \begin{minipage}{\textwidth}
        \begin{center}
        \captionof{table}{
            Cross-dataset generalization results on Mip-NeRF~360 with 32, 64, and 128 input views.
        }
        \label{tab:mipnerf_main}
    \scriptsize
    \setlength{\tabcolsep}{3pt}
    \resizebox{\textwidth}{!}{%
    \begin{tabular}{lcc|cccc|cccc|cccc}

    \toprule

        \multirow{2}{*}{Method}
        & \multirow{2}{*}{$p$}
        & \multirow{2}{*}{$k$}
        & \multicolumn{4}{c|}{32 views}
        & \multicolumn{4}{c|}{64 views}
        & \multicolumn{4}{c}{128 views} \\
        \cmidrule(lr){4-7}
        \cmidrule(lr){8-11}
        \cmidrule(lr){12-15}
        & &
        & \# GS & PSNR $\uparrow$ & SSIM $\uparrow$ & LPIPS $\downarrow$
        & \# GS & PSNR $\uparrow$ & SSIM $\uparrow$ & LPIPS $\downarrow$
        & \# GS & PSNR $\uparrow$ & SSIM $\uparrow$ & LPIPS $\downarrow$ \\

    \midrule

        YoNoSplat
        & \checkmark & \checkmark
        & 1.6M & 20.735 & 0.591 & 0.275
        & 3.2M & 21.270 & 0.620 & 0.249
        & 6.4M & 20.214 & 0.581 & 0.282 \\

        2Xplat
        & \checkmark & \checkmark
        & 1.6M & 20.766 & 0.600 & 0.333
        & 3.2M & 21.284 & 0.628 & 0.315
        & 6.4M & 21.250 & 0.646 & 0.307 \\

        ZipSplat
        & \checkmark & \checkmark
        & 332K & 20.770 & 0.562 & 0.284
        & 664K & 20.902 & 0.572 & 0.271
        & 1.3M & 20.868 & 0.571 & 0.267 \\


        Ours
        & \checkmark & \checkmark
        & 401K & \textbf{23.257} & \textbf{0.707} & \textbf{0.236}
        & 803K & \textbf{24.049} & \textbf{0.746} & \textbf{0.204}
        & 1.6M & \textbf{24.576} & \textbf{0.772} & \textbf{0.186} \\

        Ours$_\mathrm{prune}$
        & \checkmark & \checkmark
        & 332K & 23.235 & 0.705 & 0.246
        & 664K & 24.038 & 0.745 & 0.210
        & 1.3M & 24.572 & \textbf{0.772} & 0.187 \\





    \midrule

        YoNoSplat
        & &
        & 1.6M & 17.056 & 0.363 & 0.473
        & 3.2M & 16.960 & 0.364 & 0.490
        & 6.4M & 16.186 & 0.350 & 0.538 \\

        2Xplat
        & &
        & 1.6M & 20.400 & 0.558 & 0.357
        & 3.2M & 20.800 & 0.569 & 0.349
        & 6.4M & 20.743 & 0.577 & 0.345 \\

        ZipSplat
        & &
        & 332K & 21.576 & 0.581 & 0.298
        & 664K & 21.855 & 0.601 & 0.275
        & 1.3M & 21.974 & 0.610 & 0.268 \\

        Ours
        & &
        & 401K & \textbf{22.459} & \textbf{0.650} & \textbf{0.255}
        & 803K & \textbf{23.005} & \textbf{0.678} & \textbf{0.232}
        & 1.6M & \textbf{23.463} & \textbf{0.706} & \textbf{0.214} \\

        Ours$_\mathrm{prune}$
        & &
        & 332K & 22.446 & 0.649 & 0.265
        & 664K & 23.001 & 0.677 & 0.236
        & 1.3M & 23.462 & \textbf{0.706} & 0.215 \\





    \bottomrule

    \end{tabular}%
    }
        \end{center}
    \end{minipage}
    
    \vspace{0.5em}
    
    \begin{minipage}{0.99\textwidth}
        \begin{center}
        \includegraphics[width=\textwidth]{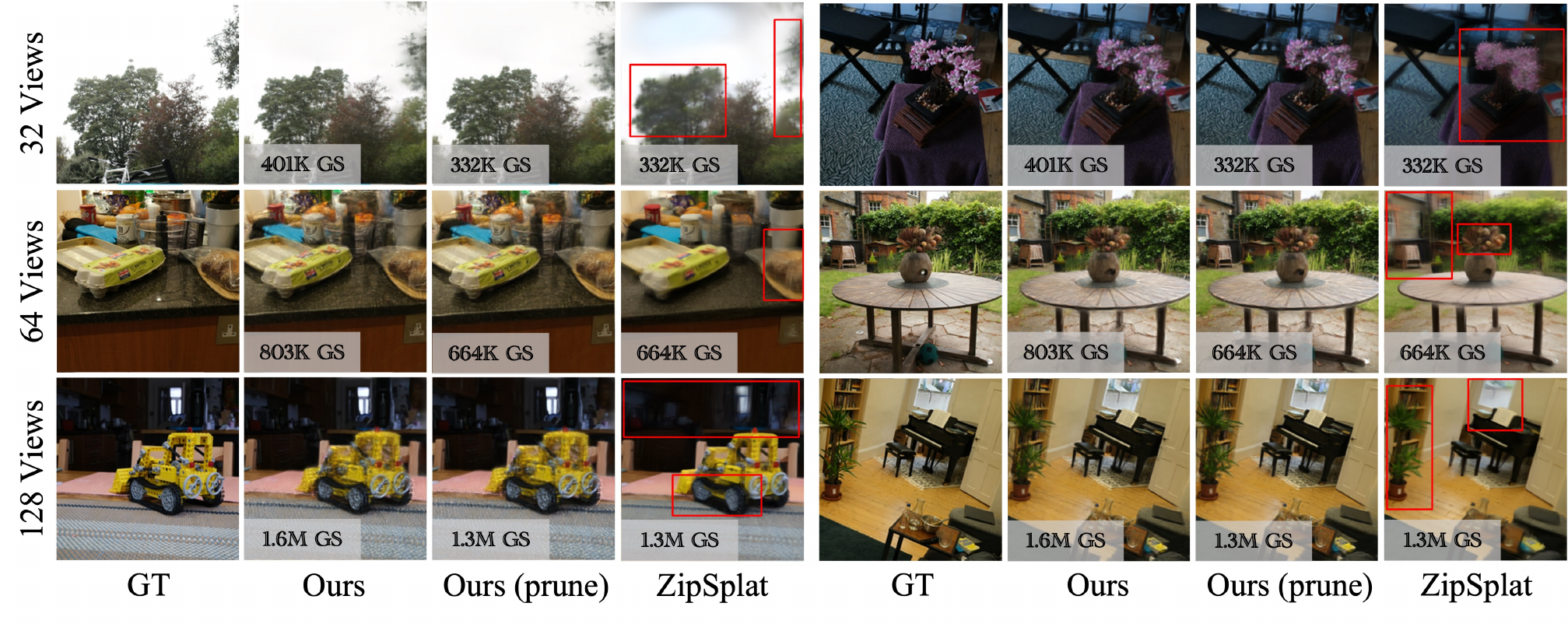}
        \captionof{figure}{
            Qualitative comparison on Mip-NeRF~360 with varying numbers of input views.
        }
        \label{fig:mip360_main_qualitative}
        \end{center}
    \end{minipage}
\end{figure}



On RE10K, as shown in Table~\ref{tab:re10k_main}, ours achieves the best performance across PSNR, SSIM, and LPIPS in the posed setting while using substantially fewer Gaussians, and it is also the best among the zero-shot methods in the pose-free setting.
Our model has only seen DL3DV, whereas the methods shown in \textcolor{gray}{gray} are trained on RE10K itself.
Although differences in training data make direct comparisons less controlled, these results nevertheless demonstrate compelling reconstruction quality and cross-dataset generalization. Qualitative comparisons are shown in Figure~\ref{fig:re10k_main_qualitative}.

On Mip-NeRF~360, presented in Table~\ref{tab:mipnerf_main}, \emph{FineSplat} achieves the highest PSNR at every view count in both settings, by 2.49 to 3.33 dB in the posed setting and 0.88 to 1.49 dB in the pose-free setting.
It also makes the best use of additional views: from 32 to 128 context views the baselines gain at most $0.5$~dB or lose ground, while ours rises by $1.32$~dB in the posed setting and $1.00$~dB in the pose-free setting, with fewer Gaussians than the pixel-aligned baselines.
We attribute this to Stage 2, which lets tokens observing the same region exchange information in 3D instead of producing redundant predictions.
Figure~\ref{fig:mip360_main_qualitative} presents the qualitative results.
\section{Analysis}
\label{sec:analysis}
We analyze how the local interactions in 3D performed by Stage~2 improve the image-space prediction of Stage~1. We describe the components of Stage~2 in the order in which information flows, and then show how they benefit from more input views.
All results in this section are posed results on DL3DV, where the context views cover a larger portion of the scene as their number increases.

\begin{figure}
    \centering
    \includegraphics[width=1.0\columnwidth]{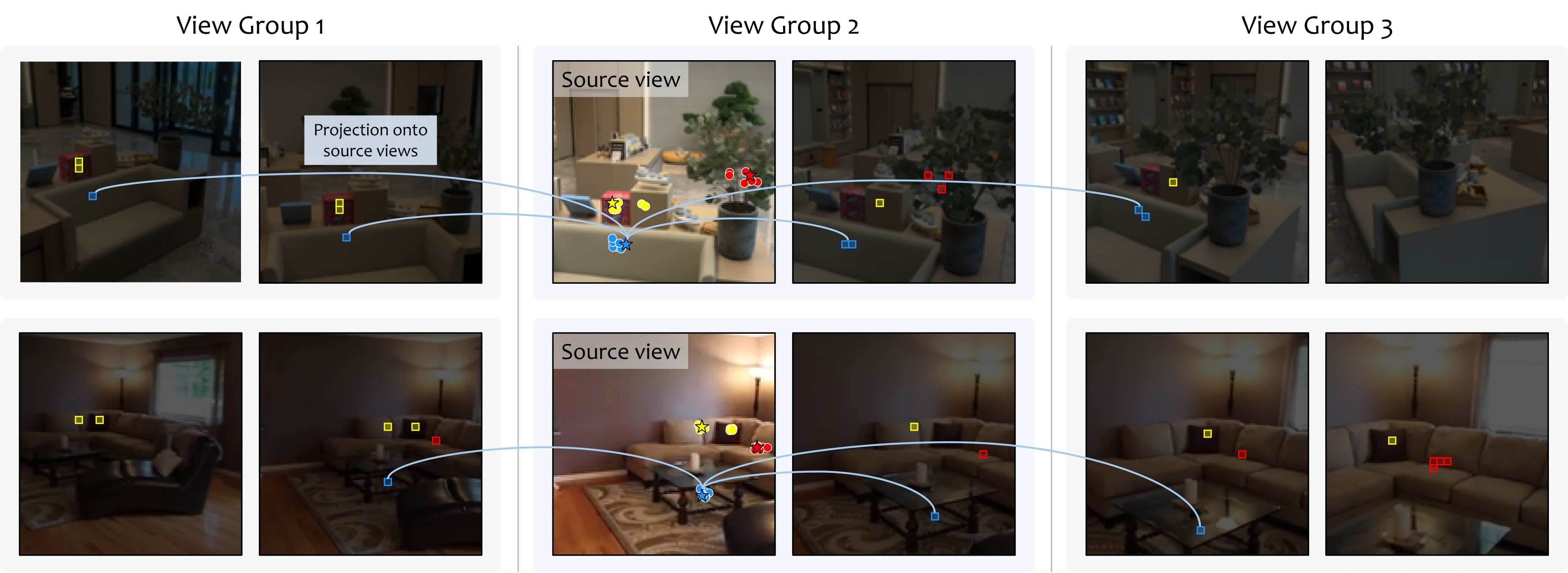}
    \caption{\textbf{Neighbors within a serialized attention window in 3D space.}
    Tokens in the same 3D spatial window share a color.
    \textit{Star}: The query tokens in its source view.
    \textit{Circle}: 3D positions of the tokens in the attention window projected onto the query's source view.
    \textit{Square}: Image patches of the tokens in the attention window in their own source views.}
    \label{fig:serialAttn}
\end{figure}

\paragraph{Cross-group interactions in 3D space.}
Stage~2 combines two error signals for each token.
The self error feature $e^{\text{self}}$ provides a direct signal from the patch where the token is predicted, while the aggregated error feature $e^{\text{agg}}$ averages the error features over all context views in which the token is visible, including views outside the view group defined in Stage~1.
Tokens then exchange these signals with their neighbors in 3D space through serialized attention~\citep{wu2024ptv3}.
Stage~2 consists of four attention blocks and each block uses a different one of the four space-filling curves, so that each token meets a different set of neighbors. 
Figure~\ref{fig:serialAttn} visualizes the window of a query token, which is a single $4{\times}4$ patch in the source view.
Its neighbors come from many different frames yet lie close to the query in 3D, showing that tokens observing the same surface can exchange information across views and view groups.

\paragraph{More Stage~2 steps and context views.}
\begin{table}[t]

    \caption{
        Rendering quality at each Stage~2 step on DL3DV with 6, 12, and 24 input views. Step~0 denotes the prediction from Stage~1. The model is trained with four steps of Stage~2, and gray rows go beyond this training setting.
    }
    \label{tab:abl_iters}

    \begin{center}
    \scriptsize
    \setlength{\tabcolsep}{3pt}
    \resizebox{0.9\textwidth}{!}{%
    \begin{tabular}{c|ccc|ccc|ccc}

    \toprule

        \multirow{2}{*}{Step}
        & \multicolumn{3}{c|}{6 views}
        & \multicolumn{3}{c|}{12 views}
        & \multicolumn{3}{c}{24 views} \\
        \cmidrule(lr){2-4}
        \cmidrule(lr){5-7}
        \cmidrule(lr){8-10}
        &
        PSNR $\uparrow$ & SSIM $\uparrow$ & LPIPS $\downarrow$
        & PSNR $\uparrow$ & SSIM $\uparrow$ & LPIPS $\downarrow$
        & PSNR $\uparrow$ & SSIM $\uparrow$ & LPIPS $\downarrow$ \\

    \midrule

        0
        & 26.109 & 0.852 & 0.133
        & 25.794 & 0.846 & 0.141
        & 25.776 & 0.850 & 0.141 \\

    \midrule

        1
        & 26.959 & 0.866 & 0.121
        & 26.741 & 0.862 & 0.127
        & 26.814 & 0.867 & 0.125 \\

        2
        & 27.201 & 0.870 & 0.117
        & 27.005 & 0.866 & 0.123
        & 27.097 & 0.872 & 0.121 \\

        3
        & 27.289 & \textbf{0.871} & 0.116
        & 27.100 & 0.867 & \textbf{0.121}
        & 27.193 & \textbf{0.873} & \textbf{0.119} \\

        4
        & 27.325 & \textbf{0.871} & \textbf{0.115}
        & \textbf{27.140} & \textbf{0.868} & \textbf{0.121}
        & \textbf{27.233} & \textbf{0.873} & \textbf{0.119} \\

    \midrule

        \rowcolor{gray!15}
        8
        & \textbf{27.333} & 0.870 & \textbf{0.115}
        & 27.137 & 0.867 & 0.122
        & 27.227 & 0.872 & 0.120 \\

        \rowcolor{gray!15}
        16
        & 27.143 & 0.865 & 0.120
        & 26.783 & 0.857 & 0.130
        & 26.796 & 0.860 & 0.131 \\

    \bottomrule

    \end{tabular}%
    }
    \end{center}

\end{table}

Table~\ref{tab:abl_iters} reports the rendering quality at each step of Stage~2.
With fine-grained patch tokens, Stage~1 alone already achieves strong performance, providing a reliable initial prediction.
However, for efficiency, its fine-grained tokens interact only within local neighborhoods of views, while interactions across all views are handled at a coarser level.
Consequently, its quality may gradually degrade since a single pass must reconstruct a larger scene and tokens observing the same surface are more often separated into different view groups.
Stage~2 compensates for this degradation, and its gain increases from 1.22~dB with 6 views to 1.46~dB with 24 views.
This trend shows that, as more views provide richer observations of each region, the 3D interactions in Stage~2 allow tokens from different view groups to combine these observations and error information.
As a result, the additional views lead to better predictions instead of being split across separate groups, and the model maintains consistently high quality across various numbers of context views.
Even a single step improves PSNR by 0.85~dB with 6 views and by 1.04~dB with 24 views, showing that 3D interactions effectively enrich Stage~1 predictions with complementary information from other views and reconstruction error feedback.
Our model is trained with four Stage~2 steps, and quality improves gradually with more steps, while going beyond the training setting yields no further gain.

\paragraph{Efficiency.}
\begin{figure}[t]
    \centering
    \includegraphics[width=0.70\textwidth]{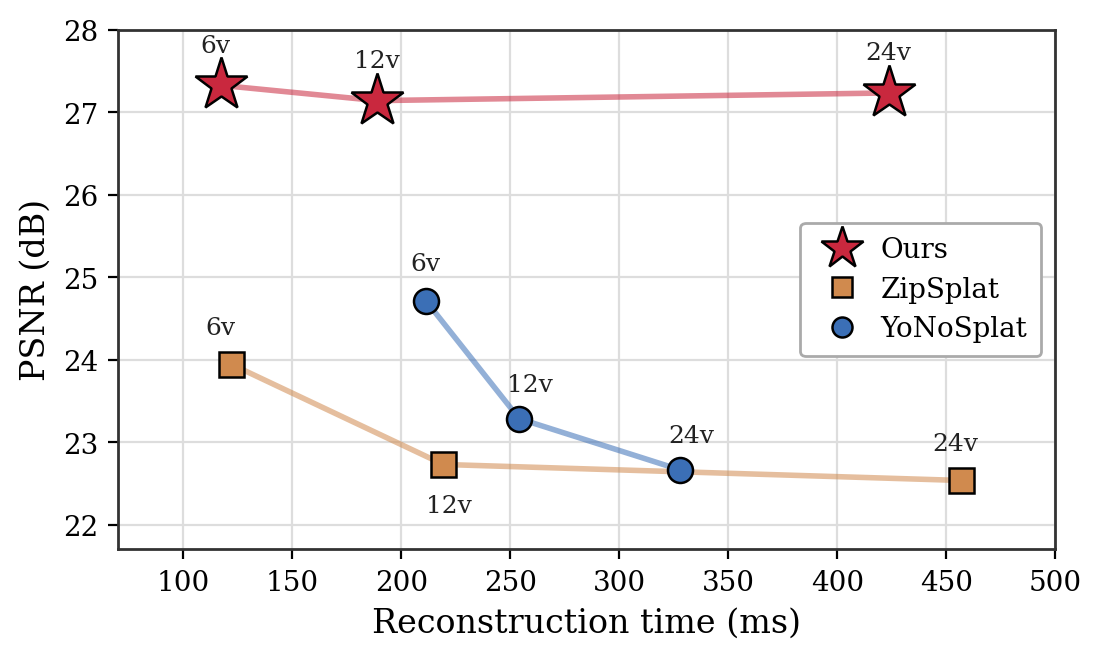}
    \caption{
        PSNR against reconstruction time on DL3DV in the posed setting, with each marker annotated by the number of context views.
    }
    \label{fig:time_psnr}
\end{figure}

Figure~\ref{fig:time_psnr} plots reconstruction time against quality for a representative pixel-aligned model, YoNoSplat, and a patch-aligned model, ZipSplat.
Both operate on $14\times14$ patches, while \emph{FineSplat} works with much finer $4\times4$ patches and remains the fastest of the three with 6 and 12 input views.
With 24 views it is still faster than ZipSplat and costs moderately more time than YoNoSplat, which produces four times as many Gaussians.
Across view counts \emph{FineSplat} returns the largest quality gain per unit of time, leading the stronger of the two baselines by $2.6$~dB with 6 input views and by $4.6$~dB with 24 views.
This efficiency comes from our dual-grouping strategy.
Instead of attending globally at every layer, Stage~1 exchanges information across neighboring views in 2D and Stage~2 across spatially nearby Gaussians in 3D.
Observations of the same surface from different views are therefore consolidated rather than left as redundant or conflicting Gaussians.
\section{Conclusion and Discussion}
We presented \emph{FineSplat}, a feed-forward 3DGS framework that preserves fine-grained visual information through high-resolution patch representations while avoiding the prohibitive cost of global multi-view attention via a \emph{dual-grouping} strategy that combines view-wise and 3D spatial locality. Together with the reconstruction feedback and compact Gaussian generation, we achieve strong novel-view synthesis quality with a small primitive budget.

While our results suggest that fine-grained multi-view interaction is an important ingredient for high-quality feed-forward reconstruction, we do not claim that fine patches are universally preferable: sufficiently large foundation models trained at scale may compensate for coarse tokenization. Nevertheless, our findings indicate that, in the multi-view transformer, revisiting token granularity and explicitly enabling fine-scale cross-view interaction can be highly effective. 
We also deliberately avoid relying on pretrained foundation-model features for Gaussian prediction, using DA3 only for camera pose estimation. In our preliminary experiments, combining such features with fine-grained tokens did not show any noticeable improvement. Exploring how coarse, semantically and geometrically rich foundation-model features can be more effectively integrated with fine-grained local representations, particularly to improve generalization, remains a promising direction for future work.



\bibliography{iclr2027_conference}
\bibliographystyle{iclr2027_conference}

\clearpage
\appendix
\section{Additional Implementation Details}
\label{sec:implementation_details}
\subsection{Architecture}
\paragraph{Camera pose estimator.}
In the pose-free setting, camera poses are predicted by a pose estimator initialized from DA3-Giant~\citep{lin2026da3}, following~\citet{jeong2026twoxplat}. It consists of 40 transformer layers with a hidden dimension of 1,536 and 24 attention heads, each with a head dimension of 64. The predicted cameras are used to condition both Stage~1 and Stage~2.

\paragraph{View-wise grouping network.}
Stage~1 is based on MVP~\citep{kang2026mvp} and adopts a 14-layer hierarchical transformer that takes $4\times4$ image patches as input. As the tokens become coarser through the layers, both the feature dimension and the range of attention increase. The first two layers operate on $4\times4$ patch tokens with a feature dimension of 256, and both use frame-wise attention within each view. The next four layers operate on $8\times8$ patch tokens with a feature dimension of 512 and consist of two pairs of a frame-wise attention layer followed by a view-wise group attention layer. For the group attention, the views are split into non-overlapping groups of two frames that are adjacent in frame index. The final eight layers operate on $16\times16$ patch tokens with a feature dimension of 1,024 and consist of four pairs of a frame-wise attention layer followed by a global attention layer across all views. All attention modules use a head dimension of 64.

\paragraph{Error features.}
Stage~2 operates on the $4\times4$ patch tokens produced by the DPT head of Stage~1. The error feature map is computed at the same $\frac{H}{4}\times\frac{W}{4}$ resolution and has 256 channels. It consists of a 128-channel error branch and a 128-channel image branch. The error branch combines the space-to-depth pixel error with the feature difference of a frozen ImageNet-pretrained ResNet-18~\citep{he2016deep}, using the conv1, layer1, and layer2 features (256 channels in total). The image branch projects the 256-dimensional finest-level features. For each token, the error feature of its source view and the mean over the views that pass the frustum and occlusion tests are concatenated and projected to 256 channels. The 1024-dimensional token is projected to 512 channels, concatenated with its error feature, and fused into a 512-dimensional point feature.

\paragraph{3D spatial grouping network.}
Stage~2 follows Point Transformer V3~\citep{wu2024ptv3} for 3D serialization and grouping, and consists of four pre-norm transformer blocks with a feature dimension of 512 and an MLP ratio of 4. Self-attention is computed in a 256-dimensional subspace with 16 heads of dimension 16. The GS tokens lifted into 3D are voxelized on a $384^3$ grid over the scene bounding box and serialized along four space-filling curves, namely Z-order, Hilbert, and their transposed variants, with one curve assigned to each block. Each block then computes local attention within groups of $S=8$ consecutive points. A final linear layer maps the output back to a 1024-dimensional residual that updates each GS token. Stage~2 has 13.2M trainable parameters and is applied with $T=4$ steps, sharing the same weights across all steps.

\subsection{Training}
\textit{FineSplat} is trained in three phases. Stage~1 is trained first, Stage~2 is then trained on top of the frozen Stage~1, and finally the whole model is fine-tuned end-to-end. In all phases, each training scene samples its number of input views from $\{6, 12, 24, 32\}$ with $\{6, 6, 6, 1\}$ target views, respectively, and the per-GPU batch is $\{4, 2, 1, 1\}$ scenes. All models are trained on NVIDIA B200 GPUs.

\paragraph{Stage~1.}
Stage~1 is first trained with pixel-aligned Gaussian prediction for 120K iterations. The pixel-aligned decoder is then replaced with our patch-aligned Gaussian decoder, and the model is fine-tuned on 8 GPUs for an additional 80K iterations. The remaining setup, including the camera losses on the predicted poses and intrinsics, follows~\citet{jeong2026twoxplat}.

\paragraph{Stage~2.}
With Stage~1 frozen, we train Stage~2 using ground-truth camera poses and intrinsics, so that it learns to correct the Gaussians independently of pose errors.
The output at every step is supervised.
Let $\hat{I}^{(i)}$ denote the target views rendered from the Gaussians after the $i$-th step, and $I$ the corresponding ground-truth images.
The training loss is averaged over all $T=4$ steps,
\begin{equation}
    \mathcal{L} = \frac{1}{T} \sum_{i=1}^{T} \Big( \mathcal{L}_{2}\big(\hat{I}^{(i)}, I\big) + \lambda\, \mathcal{L}_{\text{perc}}\big(\hat{I}^{(i)}, I\big) \Big),
    \label{eq:stage2_loss}
\end{equation}
where $\mathcal{L}_{2}$ is the mean squared error, $\mathcal{L}_{\text{perc}}$ is a VGG-19 perceptual loss, and $\lambda = 0.2$ as in Stage~1.
The tokens are detached between steps, so that each step learns to improve the state it receives.
Stage~2 is trained for 20K iterations on 4 GPUs.

\paragraph{End-to-end fine-tuning.}
Finally, we unfreeze all parameters, including the camera pose estimator and the Gaussian head, and fine-tune the full model initialized from the trained Stage~1 and Stage~2. This phase uses the fully pose-free setting, where both context and target cameras are predicted and no camera loss is used.
We additionally supervise the Stage-1 Gaussians, so the loss of Eq.~\ref{eq:stage2_loss} is averaged over steps $i = 0, \dots, T$. We fine-tune for 10K iterations on 8 GPUs.

\paragraph{Optimization.}
All stages use AdamW with $\beta_1=0.9$, $\beta_2=0.95$, and a weight decay of $0.05$ applied to multi-dimensional parameters only, gradient clipping at a maximum norm of $1.0$, and bfloat16 mixed precision.
Stage~1 uses a learning rate of $2\times10^{-5}$ without warm-up, decayed to zero with a cosine schedule. For Stage~2, we use a learning rate of $1\times10^{-4}$ with a 200-iteration linear warm-up, keep it
constant until 80\% of training, and decay it to zero with a cosine schedule over the remaining 20\%. For end-to-end fine-tuning, we use a single learning rate of $2\times10^{-5}$ for all parameters with a
200-iteration linear warm-up, followed by cosine decay to zero.

\subsection{Scene Scale Normalization}
\label{sec:scene_scale}
Feed-forward 3DGS methods predict geometry in a world frame whose origin, orientation and scale are not fixed by the input images, and each method adopts a convention to define it. The scene scale can be derived either from the context camera poses alone or from estimated depth. YoNoSplat follows the former: it places the origin at the first context camera and divides all translations by the maximum pairwise distance between context camera centers. ZipSplat follows the latter: it unprojects context depth maps pre-extracted with DA3 into 3D and divides translations by the median distance of these points to the first context camera.

We derive the scale from the context camera poses alone, but in a frame centered on the context cameras rather than on the first view. Let $c_n$ and $R_n$ be the center and camera-to-world rotation of context camera $n$. For $N$ input views, we define a mean context camera with center $\bar{c}=\frac{1}{N}\sum_n c_n$ and rotation
$\bar{R}=[\,d\times f,\ d,\ f\,]$, where $f$ is the normalized mean viewing direction $R_n e_z$
and $d$ is the mean image-down direction $R_n e_y$ orthogonalized against $f$. All camera centers are then expressed in this frame and rescaled as follows:
\begin{equation}
  \hat{c}_n = \frac{\bar{R}^\top(c_n-\bar{c})}{s},
  \qquad
  s = \max_{m}\,\big\lVert \bar{R}^\top(c_m-\bar{c})\big\rVert_\infty ,
\end{equation}
where $m$ ranges over the context cameras, so that all context camera centers lie within $[-1,1]^3$ and rotations become $\bar{R}^\top R_n$. The transform is computed from the context views only and applied unchanged to the target views for both ground-truth pose in the posed setting and DA3-predicted poses in the pose-free setting.

\subsection{Occlusion Test for Error Feature Aggregation}
To build $e^{\text{agg}}$, each token gathers error features only from the views that actually observe its surface.
If another surface lies in front of the token in a view, the error at the projected pixel belongs to that occluding surface and should not be assigned to the token.
We therefore select views with a frustum test and a depth test before averaging.
Both tests use the same context cameras that place the Gaussians, which are the ground-truth cameras in the posed setting and the predicted cameras in the pose-free setting.

\paragraph{Frustum test.}
Each token $j$ is represented by $p_j$, the mean position of its four Gaussians (Eq.~\ref{eq:token_lifting}).
For every context view $n$ we transform $p_j$ into the camera frame, which gives depth $d_{jn}$.
We then project it with the intrinsics to a pixel $u_{jn}=\pi_n(p_j)=(u,v)$.
The view passes the test if the point lies in front of the camera ($d_{jn}>0$) and $u_{jn}$ falls inside the image ($0\le u<W$, $0\le v<H$).

\paragraph{Depth test.}
For each view that passes the frustum test, we render the expected depth $D_n$ from the current Gaussians.
This is the same render that produces the error map $E_n$, so it is refreshed at every step.
We sample $D_n$ bilinearly at $u_{jn}$ and keep the view if
\begin{equation}
  d_{jn} \le (1+\epsilon)\, D_n(u_{jn}), \qquad \epsilon = 0.15 .
\end{equation}
In words, the token may be at most $15\%$ behind the rendered surface.
We use a relative tolerance for two reasons.
First, $p_j$ is the mean of four Gaussians rather than a point on the surface.
Second, the expected depth is smoothed at depth discontinuities.
A relative tolerance absorbs both effects at every scene scale.
Pixels that no Gaussian covers have no occluder, and the view is kept.

\begin{figure}[t]
    \centering
    \includegraphics[width=0.9\linewidth]{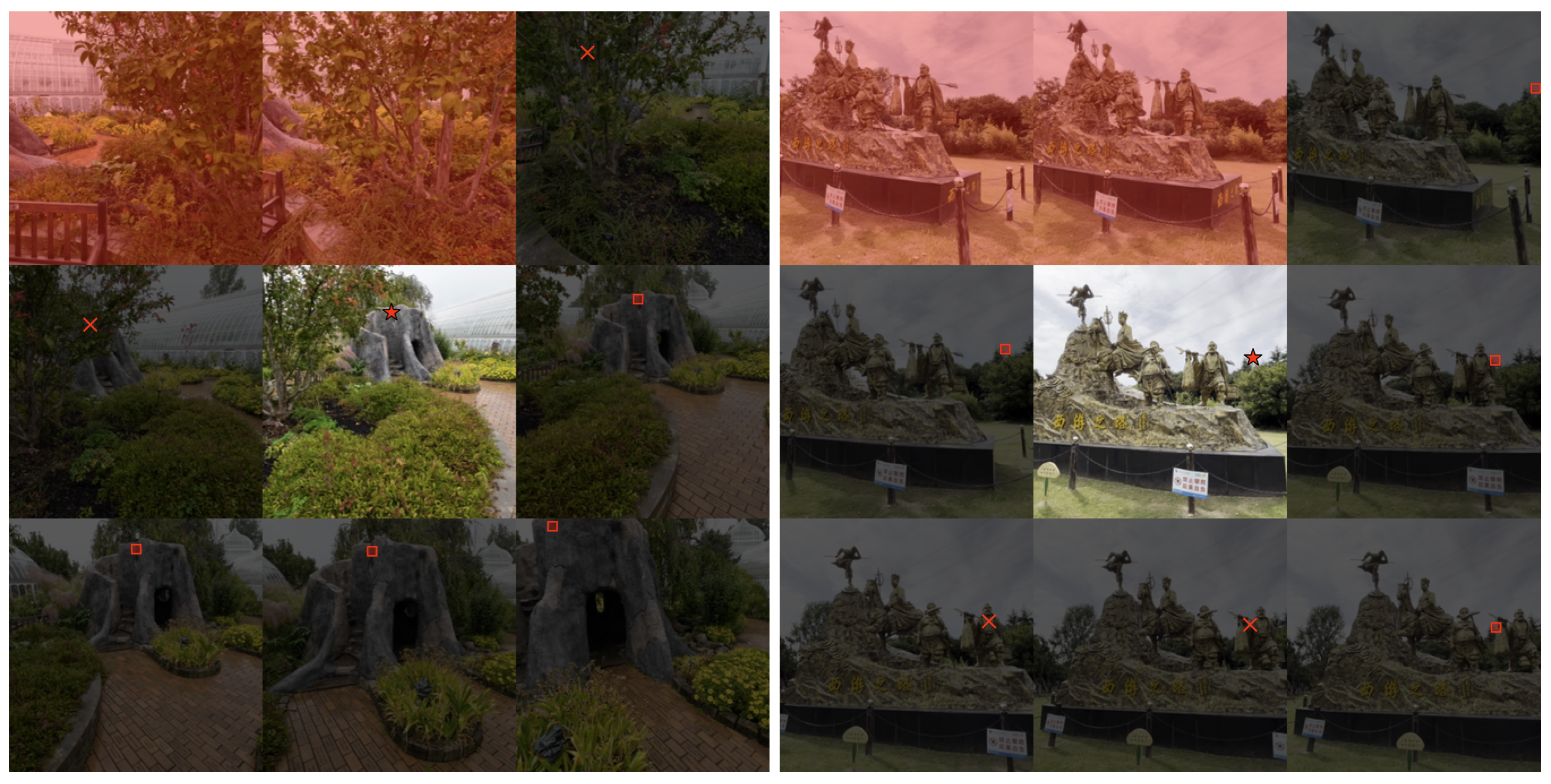}
    \caption{\textbf{View selection for $e^{\text{agg}}$.}
    The query token in the source view (center) is projected into the other context views.
    \textit{Star}: query token in its source view.
    \textit{Square}: projected location in a view that passes both the frustum and depth tests and is included in $e^{\text{agg}}$.
    \textit{Cross}: projected location in a view that fails the depth test.
    \textit{Red-tinted view}: view that fails the frustum test.}
    \label{fig:app_depth_test}
\end{figure}

Examples of views that fail the frustum test or the depth-based occlusion test are shown in Figure~\ref{fig:app_depth_test}.

\section{Evaluation Details}
\label{sec:eval_details}
\subsection{Evaluation-Time Pose Alignment}
Pose-free reconstruction is inherently ambiguous, as multiple 3D scenes and camera configurations can explain the same unposed inputs. Thus, the reconstructed scene may not be perfectly aligned with the ground-truth evaluation frame. Following prior pose-free protocols~\citep{wang2021nerf, fan2024instantsplat, ye2025noposplat}, we freeze the reconstructed 3D Gaussians and optimize each target camera pose for 200 iterations using an $\mathcal{L}_2$ loss. This evaluation-time pose alignment (EPA) is used only for pose-free settings. When ground-truth camera poses are available, we directly use them without additional alignment, following YoNoSplat~\citep{ye2026yonosplat}.

\subsection{Contribution-Based Pruning}
To evaluate a more compact representation, we prune the Gaussians predicted by our model without any additional optimization.
For each Gaussian $i$, we measure how much it contributes to the rendered context views.
Under alpha compositing, a pixel ${u}$ of view $n$ receives the blending weight $w_{i,n}({u}) = T_{i,n}({u})\,\alpha_{i,n}({u})$ from Gaussian $i$, where $\alpha_{i,n}$ is its opacity-weighted 2D footprint and $T_{i,n}$ is the transmittance accumulated in front of it.
We define the contribution score as this blending weight summed over all pixels and averaged over the $N$ context views,
\begin{equation}
    s_i = \frac{1}{N} \sum_{n=1}^{N} \sum_{{u} \in \Omega_n} T_{i,n}({u})\,\alpha_{i,n}({u}),
\end{equation}
where $\Omega_n$ denotes the set of pixels in view $n$.
This score does not depend on the color of the Gaussian and accounts for occlusion.
In practice, we obtain $s_i$ with a single backward pass per view, as the gradient of the summed rendering with respect to a unit scalar color assigned to every Gaussian.
The scores are computed once per scene using only the context images and the cameras used for evaluation, which are ground-truth or predicted depending on the setting.
We then keep the $K = k \cdot N$ Gaussians with the highest scores, where we set $k = 10{,}368$ Gaussians per view, and discard the rest.
Only the remaining Gaussians are used for rendering the target views and, in the pose-free setting, for evaluation-time pose alignment.

\subsection{Evaluation Resolution and Resize Filter}
\label{sec:eval_reslution}

\begin{figure}[t]
    \centering
    \includegraphics[width=\textwidth]{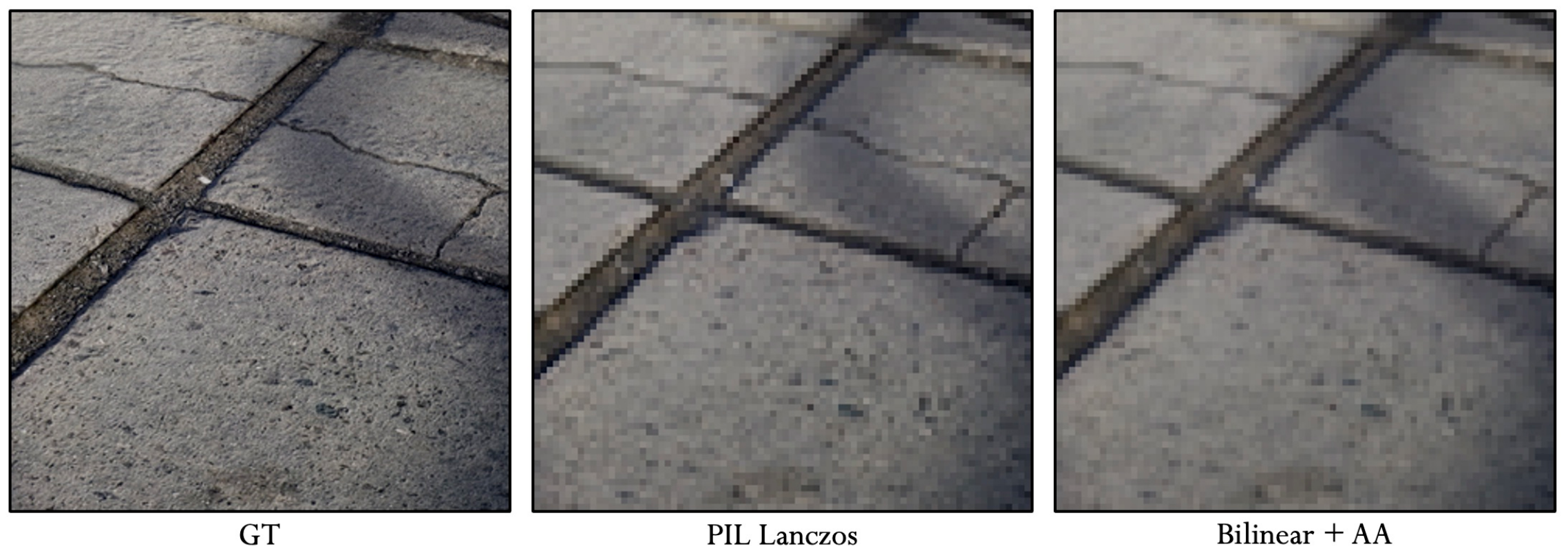}
    \caption{\textbf{Comparison of resizing filters.}
    From left to right, an original high-resolution image and its downsampled versions using PIL Lanczos and bilinear interpolation with anti-aliasing (Bilinear+AA).
    Lanczos better preserves high-frequency details such as the thin cracks and surface texture, whereas Bilinear+AA produces a smoother image in which these details are partly lost.}
    \label{fig:resize_filter_ablation}
\end{figure}

Images in the datasets are downsampled from their original resolution for both training and evaluation, and the reported metrics can vary noticeably depending on the image resolution and the resizing filter used in this process.
As shown in Figure~\ref{fig:resize_filter_ablation}, the same image yields visibly different results under different filters.
In the main paper, each method performs Gaussian reconstruction and rendering at its native training resolution, which is $256\times448$ for DepthSplat, $224\times224$ for C3G, YoNoSplat, MVP, NoDrift3R, and 2Xplat, $252\times252$ for ZipSplat, and $448\times448$ for AnySplat.
Only the final rendered and ground-truth target images are then resized to a common resolution of $224\times224$ for metric computation, following the evaluation protocol of~\citet{ye2026yonosplat}.
Since our method operates natively at $224\times224$, it uses the smallest input resolution among the compared methods, so this protocol gives it no advantage from a larger input resolution or field of view.
For this resizing, we use PIL Lanczos as the default filter, following the protocol adopted by all baselines~\citep{chen2024mvsplat, xu2025depthsplat, lin2026da3, ye2025noposplat, jiang2025anysplat, an2025c3g, ye2026yonosplat}, except ZipSplat~\citep{veicht2026zipsplat}.

\begin{table}[t]

    \caption{
        Novel view synthesis results on DL3DV with 6, 12, and 24 input views at $252\times252$ resolution, evaluated on images downsampled with bilinear interpolation with anti-aliasing (Bilinear+AA).       
    }
    \label{tab:dl3dv_main_zipsplat_setting}

    \begin{center}
    \scriptsize
    \setlength{\tabcolsep}{3pt}
    \resizebox{\textwidth}{!}{%
    \begin{tabular}{lcc|cccc|cccc|cccc}

    \toprule

        \multirow{2}{*}{Method}
        & \multirow{2}{*}{$p$}
        & \multirow{2}{*}{$k$}
        & \multicolumn{4}{c|}{6 views}
        & \multicolumn{4}{c|}{12 views}
        & \multicolumn{4}{c}{24 views} \\
        \cmidrule(lr){4-7}
        \cmidrule(lr){8-11}
        \cmidrule(lr){12-15}
        &
        &
        &
        \# GS & PSNR $\uparrow$ & SSIM $\uparrow$ & LPIPS $\downarrow$
        & \# GS & PSNR $\uparrow$ & SSIM $\uparrow$ & LPIPS $\downarrow$
        & \# GS & PSNR $\uparrow$ & SSIM $\uparrow$ & LPIPS $\downarrow$ \\

    \midrule


        DepthSplat
        & \checkmark & \checkmark
        & 393K & 24.63 & 0.837 & 0.142
        & 786K & 22.64 & 0.793 & 0.187
        & 1.6M & 20.60 & 0.734 & 0.248 \\

        DA3
        & \checkmark & \checkmark
        & 1.5M & 23.99 & 0.805 & 0.158
        & 3.0M & 22.84 & 0.758 & 0.190
        & 6.1M & 21.70 & 0.710 & 0.230 \\

        ZipSplat
        & \checkmark & \checkmark
        & 62K & 24.75 & 0.790 & 0.174
        & 124K & 23.40 & 0.735 & 0.204
        & 249K & 23.20 & 0.732 & 0.206 \\

        
        Ours 
        & \checkmark & \checkmark
        & 95K & \textbf{28.33} & \textbf{0.886} & \textbf{0.111}
        & 191K & \textbf{28.15} & \textbf{0.883} & \textbf{0.117}
        & 381K & \textbf{28.25} & \textbf{0.889} & \textbf{0.116} \\

        Ours$_\mathrm{prune}$
        & \checkmark & \checkmark
        & 62K & 27.33 & 0.868 & 0.138
        & 124K & 27.31 & 0.867 & 0.141
        & 249K & 27.64 & 0.876 & 0.134 \\




        
    \midrule

        AnySplat
        & &
        & 951K & 21.70 & 0.725 & 0.187
        & 1.8M & 21.01 & 0.687 & 0.220
        & 3.2M & 20.74 & 0.669 & 0.236 \\

        C3G
        & &
        & 2K & 18.70 & 0.492 & 0.409
        & 2K & 16.50 & 0.421 & 0.534
        & 2K & 15.17 & 0.376 & 0.583 \\

        DA3
        & &
        & 1.5M & 23.77 & 0.795 & 0.165
        & 3.0M & 22.38 & 0.736 & 0.208
        & 6.1M & 21.69 & 0.711 & 0.229 \\

        YoNoSplat
        & &
        & 301K & 24.10 & 0.783 & 0.160
        & 602K & 22.73 & 0.736 & 0.200
        & 1.2M & 22.01 & 0.710 & 0.223 \\

        ZipSplat
        & &
        & 62K & 25.24 & 0.804 & 0.172
        & 124K & 24.27 & 0.767 & 0.197
        & 249K & 24.14 & 0.768 & 0.198 \\

        Ours 
        & &
        & 95K & \textbf{28.11} & \textbf{0.879} & \textbf{0.113}
        & 191K & \textbf{27.89} & \textbf{0.875} & \textbf{0.120}
        & 381K & \textbf{27.92} & \textbf{0.877} & \textbf{0.122} \\

        Ours$_\mathrm{prune}$
        & &
        & 62K & 27.18 & 0.862 & 0.139
        & 124K & 27.13 & 0.860 & 0.143
        & 249K & 27.37 & 0.865 & 0.139 \\
        
    \bottomrule
    \end{tabular}%
    }
    \end{center}
\end{table}

We additionally report results at a different resolution and with a different resizing filter, following the protocol of ZipSplat~\citep{veicht2026zipsplat}.
Specifically, we fine-tune our model at $252\times252$ and perform both reconstruction and rendering at this resolution, where both the input images and the ground-truth target images are resized with bilinear interpolation with anti-aliasing (Bilinear+AA).
Only ZipSplat is also trained on images resized with Bilinear+AA, while our model and the other methods are trained on images resized with Lanczos.
Since the evaluation protocol is identical, the results of the baselines in Table~\ref{tab:dl3dv_main_zipsplat_setting} are taken from~\citet{veicht2026zipsplat}.
\section{Additional Analysis}
\label{sec:additional_analysis}
\subsection{View Grouping and the Choice of Group Size}
\begin{table}[t]

    \caption{
        Ablation of the view-group size on DL3DV with 6, 12, and 24 input views.
        The model is trained with a group size of 2, and only the group size of the view-wise grouping is changed at inference.
        Stage~1 denotes the initial prediction, and Stage~2 the output of the full model.
    }
    \label{tab:abl_viewgroup}

    \begin{center}
    \scriptsize
    \setlength{\tabcolsep}{3pt}
    \resizebox{\textwidth}{!}{%
    \begin{tabular}{cc|ccc|ccc|ccc}

        \toprule

        \multirow{2}{*}[-2pt]{Stage}
        & \multirow{2}{*}[-2pt]{\makecell{View-group\\size}}
        & \multicolumn{3}{c|}{6 views}
        & \multicolumn{3}{c|}{12 views}
        & \multicolumn{3}{c}{24 views} \\
        \cmidrule(lr){3-5} \cmidrule(lr){6-8} \cmidrule(lr){9-11}
        & & PSNR $\uparrow$ & SSIM $\uparrow$ & LPIPS $\downarrow$
          & PSNR $\uparrow$ & SSIM $\uparrow$ & LPIPS $\downarrow$
          & PSNR $\uparrow$ & SSIM $\uparrow$ & LPIPS $\downarrow$ \\

    \midrule

        & 1
        & 21.928 & 0.780 & 0.183
        & 23.165 & 0.800 & 0.172
        & 23.851 & 0.817 & 0.163 \\

        & 2
        & \textbf{26.109} & \textbf{0.852} & \textbf{0.133}
        & \textbf{25.794} & \textbf{0.846} & \textbf{0.141}
        & \textbf{25.776} & \textbf{0.850} & \textbf{0.141} \\

        & 4
        & -- & -- & --
        & 25.456 & 0.839 & 0.147
        & 25.447 & 0.844 & 0.146 \\

        & 6
        & 25.491 & 0.834 & 0.147
        & 25.074 & 0.829 & 0.155
        & 25.001 & 0.835 & 0.153 \\

        & 8
        & -- & -- & --
        & -- & -- & --
        & 24.628 & 0.827 & 0.160 \\

        & 12
        & -- & -- & --
        & 24.307 & 0.805 & 0.174
        & 24.063 & 0.812 & 0.173 \\

        \multirow{-7}{*}{Stage~1}
        & 24
        & -- & -- & --
        & -- & -- & --
        & 23.247 & 0.784 & 0.197 \\

    \midrule

        & 1
        & 26.164 & 0.846 & 0.139
        & 26.446 & 0.851 & 0.138
        & 26.691 & 0.861 & 0.133 \\

        & 2
        & \textbf{27.325} & \textbf{0.871} & \textbf{0.115}
        & \textbf{27.140} & \textbf{0.868} & \textbf{0.121}
        & \textbf{27.233} & \textbf{0.873} & \textbf{0.119} \\

        & 4
        & -- & -- & --
        & 26.882 & 0.862 & 0.125
        & 27.053 & 0.869 & 0.122 \\

        & 6
        & 26.835 & 0.858 & 0.125
        & 26.617 & 0.856 & 0.130
        & 26.812 & 0.864 & 0.126 \\

        & 8
        & -- & -- & --
        & -- & -- & --
        & 26.590 & 0.858 & 0.131 \\

        & 12
        & -- & -- & --
        & 25.997 & 0.839 & 0.144
        & 26.204 & 0.849 & 0.139 \\

        \multirow{-7}{*}{Stage~2}
        & 24
        & -- & -- & --
        & -- & -- & --
        & 25.465 & 0.828 & 0.157 \\

    \bottomrule

    \end{tabular}%
    }
    \end{center}

\end{table}








Fine-grained tokens make multi-view attention expensive, and Stage~1 therefore restricts it to small groups of views.
This grouping is introduced for efficiency, but Table~\ref{tab:abl_viewgroup}, which varies the group size at inference while the model is trained with a group size of 2, shows that it is not a compromise.
Removing the grouping altogether costs the most: with a group size of 1 each view is processed on its own, fine-grained cross-view interaction disappears, and PSNR at Stage~1 drops by 4.18~dB with 6 views and by 1.93~dB with 24 views.
Simply enlarging the group does not improve quality.
Performance decreases monotonically as the group grows, and with 24 views a single group spanning all views is 2.53~dB below the default.
A group size of 2 is the best setting at every view count and at both stages.
Attending over all views at this granularity is therefore not simply better, and its cost grows quadratically with the number of tokens in a group.
Small groups of views keep fine-grained attention affordable and, at the same time, give the operating point at which it works best.

Stage~2 further relaxes the choice of the group size.
With a group size of 1, the gap to the default shrinks from 4.18~dB to 1.16~dB with 6 views and from 1.93~dB to 0.54~dB with 24 views.
A single global group with 24 views narrows from 2.53~dB to 1.77~dB.
The interactions in 3D recover much of the cross-view information that Stage~1 missed, so the model does not depend on the grouping being exactly right.

\subsection{Error Feature Composition}
\begin{table}[t]

    \caption{
        Ablation on the error feature on DL3DV with 6, 12, and 24 input views.
    }
    \label{tab:abl_err}

    \begin{center}    
    \scriptsize
    \setlength{\tabcolsep}{3pt}
    \resizebox{\textwidth}{!}{%
    \begin{tabular}{cc|ccc|ccc|ccc}

    \toprule

        \multirow{2}{*}{$e^{\text{self}}$}
        & \multirow{2}{*}{$e^{\text{agg}}$}
        & \multicolumn{3}{c|}{6 views}
        & \multicolumn{3}{c|}{12 views}
        & \multicolumn{3}{c}{24 views} \\
        \cmidrule(lr){3-5}
        \cmidrule(lr){6-8}
        \cmidrule(lr){9-11}
        &
        & PSNR $\uparrow$ & SSIM $\uparrow$ & LPIPS $\downarrow$
        & PSNR $\uparrow$ & SSIM $\uparrow$ & LPIPS $\downarrow$
        & PSNR $\uparrow$ & SSIM $\uparrow$ & LPIPS $\downarrow$ \\

    \midrule


                   & \checkmark
        & 26.890 & 0.864 & 0.121
        & 26.586 & 0.858 & 0.129
        & 26.611 & 0.863 & 0.127 \\

        \checkmark &
        & 27.236 & 0.870 & 0.116
        & 27.000 & 0.865 & 0.123
        & 27.063 & 0.870 & 0.122 \\

        \checkmark & \checkmark
        & 27.325 & 0.871 & 0.115
        & 27.140 & 0.868 & 0.121
        & 27.233 & 0.873 & 0.119 \\
        
    \bottomrule

    \end{tabular}%
    }
    \end{center}
\end{table}

We examine the contribution of the two error features in Table~\ref{tab:abl_err}.
Removing either feature degrades the quality in all settings, indicating that the two features provide complementary information.
Removing $e^{\text{self}}$ causes the largest drop, from 0.44~dB with 6 views to 0.62~dB with 24 views in PSNR.
Since $e^{\text{self}}$ is sampled at the patch where the token is predicted, it provides the most direct signal of how well the token explains its own observation, and the model relies on it most heavily.
Removing $e^{\text{agg}}$ leads to a smaller but consistent drop, which grows with the number of views from 0.09~dB to 0.17~dB.
This suggests that $e^{\text{agg}}$ becomes more useful as each token is observed by more views, which we examine further below.

\subsection{Observations Beyond the View Group}
\begin{table}[t]

    \caption{
        Masking out the views outside the view group in $e^{\text{agg}}$ at inference time.
        \emph{Within group}: $e^{\text{agg}}$ averages only the views of the token's own view group in Stage~1.
        \emph{Random}: the same number of visible views is removed at random.
    }
    \label{tab:abl_maskout}

    \begin{center}
    \scriptsize
    \setlength{\tabcolsep}{4pt}
    \resizebox{\textwidth}{!}{%
    \begin{tabular}{l|ccc|ccc|ccc}

    \toprule

        \multirow{2}{*}{$e^{\text{agg}}$ views}
        & \multicolumn{3}{c|}{6 views}
        & \multicolumn{3}{c|}{12 views}
        & \multicolumn{3}{c}{24 views} \\
        \cmidrule(lr){2-4}
        \cmidrule(lr){5-7}
        \cmidrule(lr){8-10}
        & PSNR $\uparrow$ & SSIM $\uparrow$ & LPIPS $\downarrow$
        & PSNR $\uparrow$ & SSIM $\uparrow$ & LPIPS $\downarrow$
        & PSNR $\uparrow$ & SSIM $\uparrow$ & LPIPS $\downarrow$ \\

    \midrule

        All (Ours)
        & 27.325 & 0.871 & 0.115
        & 27.140 & 0.868 & 0.121
        & 27.233 & 0.873 & 0.119 \\

    \midrule

        Within Group
        & 27.099 & 0.866 & 0.118
        & 26.734 & 0.859 & 0.128
        & 26.628 & 0.860 & 0.131 \\

        Random
        & 27.141 & 0.867 & 0.118
        & 26.816 & 0.860 & 0.126
        & 26.801 & 0.863 & 0.128 \\

    \bottomrule

    \end{tabular}%
    }
    \end{center}

\end{table}

To examine whether this growing gain comes from observations outside the view group, we restrict $e^{\text{agg}}$ at inference time to the views within the token's own view group in Table~\ref{tab:abl_maskout}.
As the number of views grows, most visible views lie outside the group, from 55.5\% with 6 views to 84.0\% with 24 views.
We therefore compare this setting with a control that removes the same number of visible views at random.
Both settings discard the same number of observations, yet removing the views outside the group consistently causes a larger drop, and the difference widens with the number of views, from 0.04~dB with 6 views to 0.17~dB with 24 views.
This shows that observations from other view groups carry information that the views within the group cannot provide.
Together with the results in Table~\ref{tab:abl_iters}, these results indicate that FineSplat gathers observations beyond the view groups of Stage~1 and exchanges them among tokens that are close in 3D, which keeps its quality stable as the number of input views increases.

\subsection{Additional Cross-Dataset Generalization}
\begin{figure}[t]
    \centering
    \begin{minipage}{\textwidth}
        \begin{center}
        \captionof{table}{
            Cross-dataset generalization results on ScanNet++ with 32, 64, and 128 input views.
            Methods shown in \textcolor{gray}{gray} are trained on ScanNet++.            
        }
        \label{tab:scannetpp}
    \scriptsize
    \setlength{\tabcolsep}{3pt}
    \resizebox{\textwidth}{!}{%
    \begin{tabular}{lcc|cccc|cccc|cccc}

    \toprule

        \multirow{2}{*}{Method}
        & \multirow{2}{*}{$p$}
        & \multirow{2}{*}{$k$}
        & \multicolumn{4}{c|}{32 views}
        & \multicolumn{4}{c|}{64 views}
        & \multicolumn{4}{c}{128 views} \\
        \cmidrule(lr){4-7}
        \cmidrule(lr){8-11}
        \cmidrule(lr){12-15}
        & &
        & \# GS & PSNR $\uparrow$ & SSIM $\uparrow$ & LPIPS $\downarrow$
        & \# GS & PSNR $\uparrow$ & SSIM $\uparrow$ & LPIPS $\downarrow$
        & \# GS & PSNR $\uparrow$ & SSIM $\uparrow$ & LPIPS $\downarrow$ \\

    \midrule

        YoNoSplat
        & \checkmark & \checkmark
        & 1.6M & 17.630 & 0.644 & 0.378
        & 3.2M & 18.746 & 0.684 & 0.330
        & 6.4M & 19.123 & 0.702 & 0.316 \\

        2Xplat
        & \checkmark & \checkmark
        & 1.6M & 18.580 & 0.675 & 0.369
        & 3.2M & 20.250 & 0.729 & 0.307
        & 6.4M & 21.260 & 0.761 & 0.276 \\



        ZipSplat 
        & \checkmark & \checkmark
        & 332K & 18.661 & 0.652 & 0.340
        & 664K & 19.773 & 0.683 & 0.296
        & 1.3M & 20.187 & 0.694 & 0.279 \\



        Ours
        & \checkmark & \checkmark
        & 401K & 19.850 & \textbf{0.715} & \textbf{0.315}
        & 803K & 21.934 & 0.774 & \textbf{0.244}
        & 1.6M & 23.329 & \textbf{0.808} & \textbf{0.209} \\

        Ours$_\mathrm{prune}$
        & \checkmark & \checkmark
        & 332K & \textbf{19.851} & \textbf{0.715} & 0.319
        & 664K & \textbf{21.935} & \textbf{0.775} & 0.246
        & 1.3M & \textbf{23.330} & \textbf{0.808} & \textbf{0.209} \\





    \midrule

        \textcolor{gray}{AnySplat}
        & &
        & \textcolor{gray}{4.5M} & \textcolor{gray}{16.829} & \textcolor{gray}{0.615} & \textcolor{gray}{0.401}
        & \textcolor{gray}{8.3M} & \textcolor{gray}{18.526} & \textcolor{gray}{0.671} & \textcolor{gray}{0.344}
        & \textcolor{gray}{14.5M} & \textcolor{gray}{19.397} & \textcolor{gray}{0.697} & \textcolor{gray}{0.321} \\

        \textcolor{gray}{DA3}
        & &
        & \textcolor{gray}{8.1M} & \textcolor{gray}{18.545} & \textcolor{gray}{0.678} & \textcolor{gray}{0.363}
        & \textcolor{gray}{16.3M} & \textcolor{gray}{19.459} & \textcolor{gray}{0.712} & \textcolor{gray}{0.318}
        & \textcolor{gray}{32.5M} & \textcolor{gray}{20.274} & \textcolor{gray}{0.740} & \textcolor{gray}{0.279} \\

        YoNoSplat
        & &
        & 1.6M & 16.886 & 0.600 & 0.432
        & 3.2M & 17.368 & 0.608 & 0.413
        & 6.4M & 17.641 & 0.617 & 0.405 \\

        NoDrift3R
        & &
        & 1.6M & 17.569 & 0.620 & 0.419
        & 3.2M & 18.935 & 0.668 & 0.348
        & 6.4M & 19.714 & 0.695 & 0.317 \\

        2Xplat
        & &
        & 1.6M & 18.043 & 0.651 & 0.402
        & 3.2M & 19.639 & 0.698 & 0.342
        & 6.4M & 20.741 & 0.731 & 0.309 \\

        C3G
        & &
        & 2K & 12.776 & 0.492 & 0.657
        & 2K & 12.595 & 0.482 & 0.660
        & 2K & 12.183 & 0.465 & 0.668 \\


        ZipSplat 
        & &
        & 332K & \textbf{19.563} & \textbf{0.682} & \textbf{0.354}
        & 664K & \textbf{20.912} & \textbf{0.725} & \textbf{0.299}
        & 1.3M & 21.465 & 0.744 & 0.280 \\


        Ours
        & &
        & 401K & 18.141 & 0.647 & 0.388
        & 803K & 20.483 & 0.719 & \textbf{0.299}
        & 1.6M & 22.389 & \textbf{0.773} & \textbf{0.245} \\

        Ours$_\mathrm{prune}$
        & &
        & 332K & 18.142 & 0.648 & 0.391
        & 664K & 20.483 & 0.720 & 0.301
        & 1.3M & \textbf{22.391} & \textbf{0.773} & \textbf{0.245} \\





    \bottomrule

    \end{tabular}%
    }
        \end{center}
    \end{minipage}
    
    \vspace{0.5em}
    
    \begin{minipage}{\textwidth}
        \begin{center}
        \includegraphics[width=\textwidth]{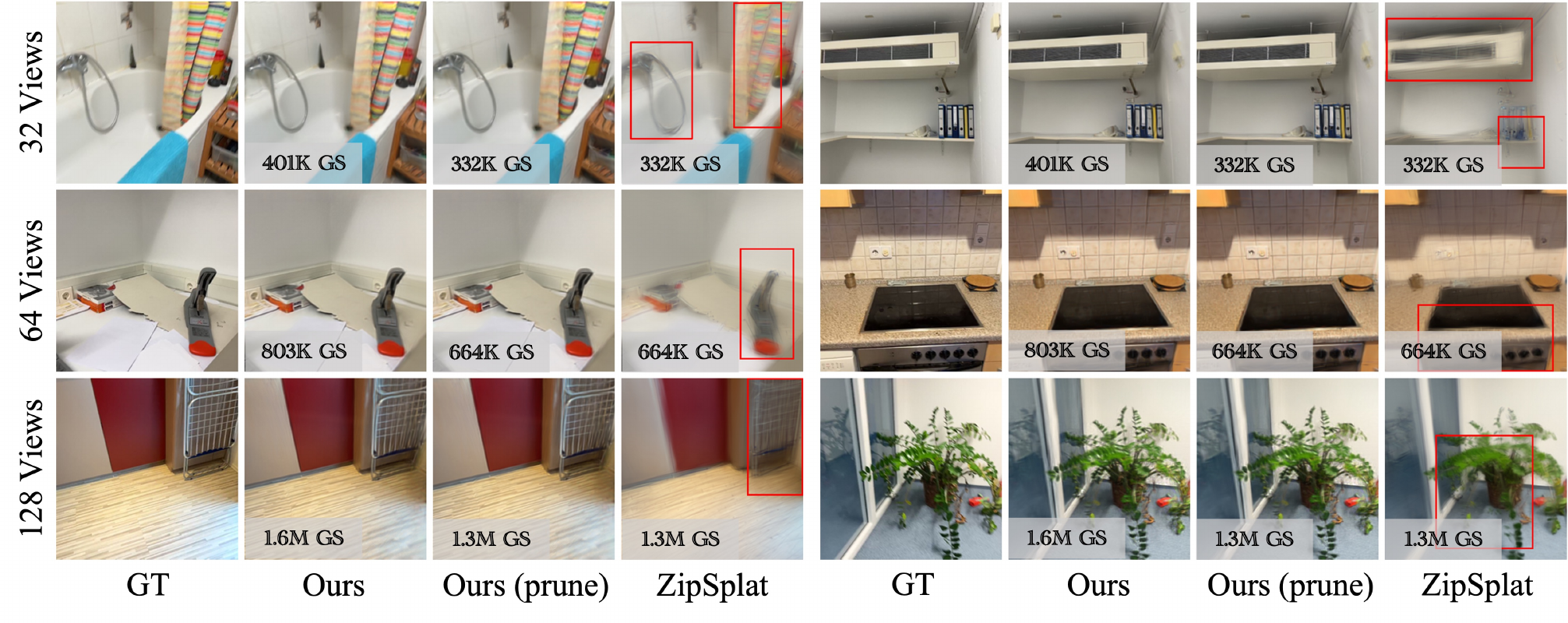}
        \captionof{figure}{
            Qualitative comparison on ScanNet++ with varying numbers of input views.
        }
        \label{fig:scannetpp_qualitative}
        \end{center}
    \end{minipage}
\end{figure}

We further evaluate \emph{FineSplat} on ScanNet++~\citep{yeshwanth2023scannet++} and report the results in Table~\ref{tab:scannetpp} and Figure~\ref{fig:scannetpp_qualitative}.
Since the model is trained on DL3DV with $\{6, 12, 24, 32\}$ views and evaluated with $\{32, 64, 128\}$ views, this setting tests generalization to both unseen scenes and unseen view counts.
In the posed setting, our method attains the highest PSNR among all methods trained on DL3DV at every view count, with the margin growing from $+1.19$~dB at 32 views to $+2.07$~dB at 128 views.
As more views are added, each surface is observed by more tokens, and Stage~2 consolidates this complementary evidence by letting each token exchange information with its 3D neighbors across views and view groups.
Consequently, at 64 and 128 views \emph{FineSplat} even surpasses methods trained directly on ScanNet++ while using far fewer Gaussians.

The pose-free setting shows the same trend from a lower starting point, as reconstruction quality depends on how well the geometry of an unfamiliar domain is recovered from estimated poses.
At 32 views, DA3, which is trained on ScanNet++, and ZipSplat lead, and at 64 views only ZipSplat remains ahead. Both rely on depth supervision, whereas \emph{FineSplat} uses photometric supervision alone and normalizes scenes with context camera poses only. Even so, the gap closes quickly, and with 128 views \emph{FineSplat} takes the lead and also exceeds the models trained directly on ScanNet++.
This is consistent with the dual-grouping design, as each additional view gives Stage~2 more evidence to anchor tokens in 3D, indicating that the benefit of 3D spatial grouping carries over to domains and view counts well beyond the training distribution.

\clearpage
\section{Additional Qualitative Results}
All results in this section follow the evaluation setting of the main paper, at a resolution of $224\times224$ with the PIL Lanczos resizing filter.

\begin{figure}[!h]
    \centering
    \includegraphics[width=\textwidth]{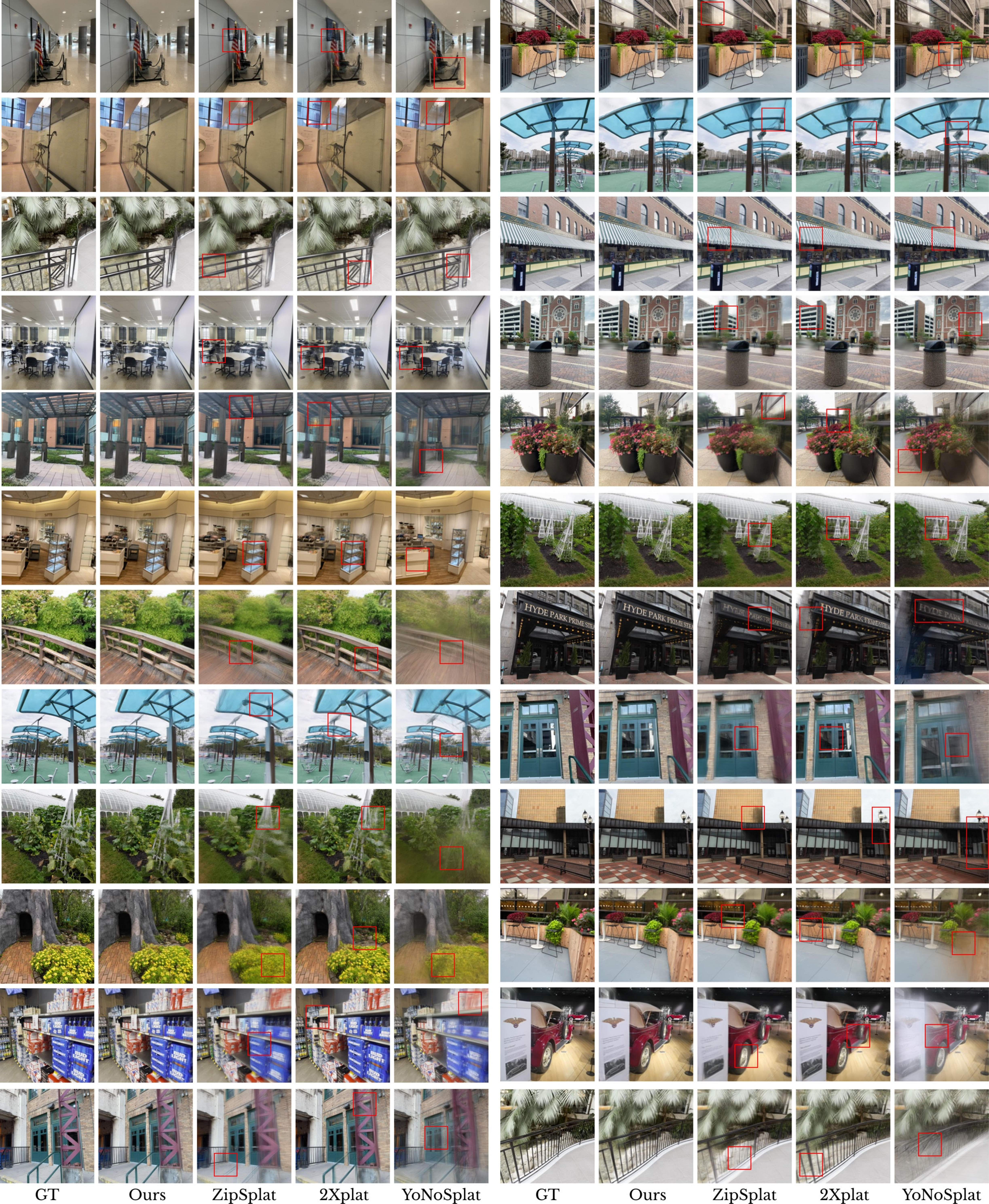}
    \caption{Qualitative comparison on DL3DV. Rows 1 to 4, 5 to 8, and 9 to 12 use 6, 12, and 24 context views, respectively.}
    \label{fig:dl3dv_ablation_qualitative}
\end{figure}

\begin{figure}[t]
    \centering
    \includegraphics[width=0.95\textwidth]{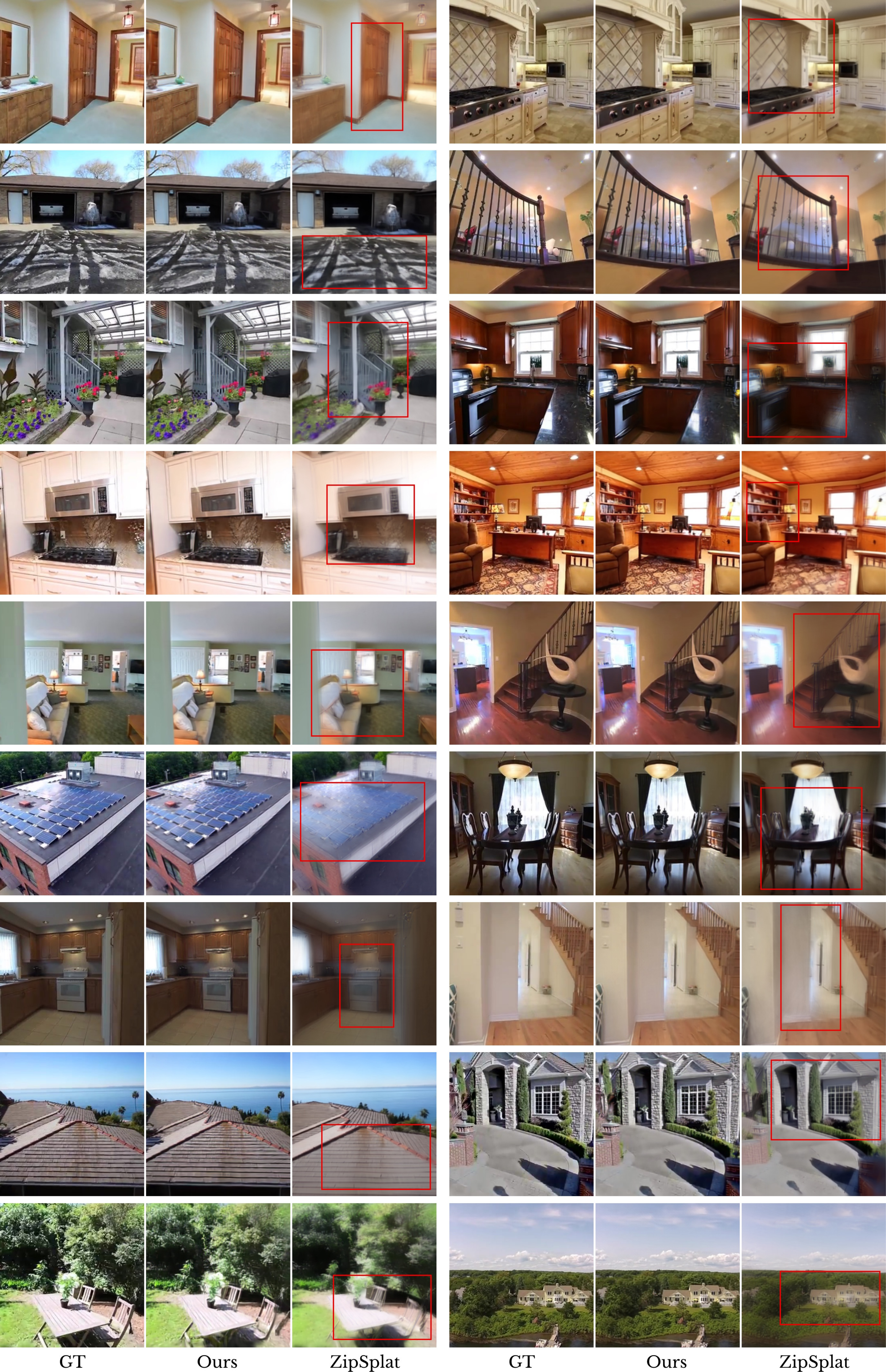}
    \caption{Qualitative comparison on RE10K with 6 input views. Our model is evaluated zero-shot without training on RE10K, whereas ZipSplat is trained on RE10K.}
    \label{fig:re10k_ablation_qualitative}
\end{figure}

\begin{figure}[t]
    \centering
    \includegraphics[width=0.95\textwidth]{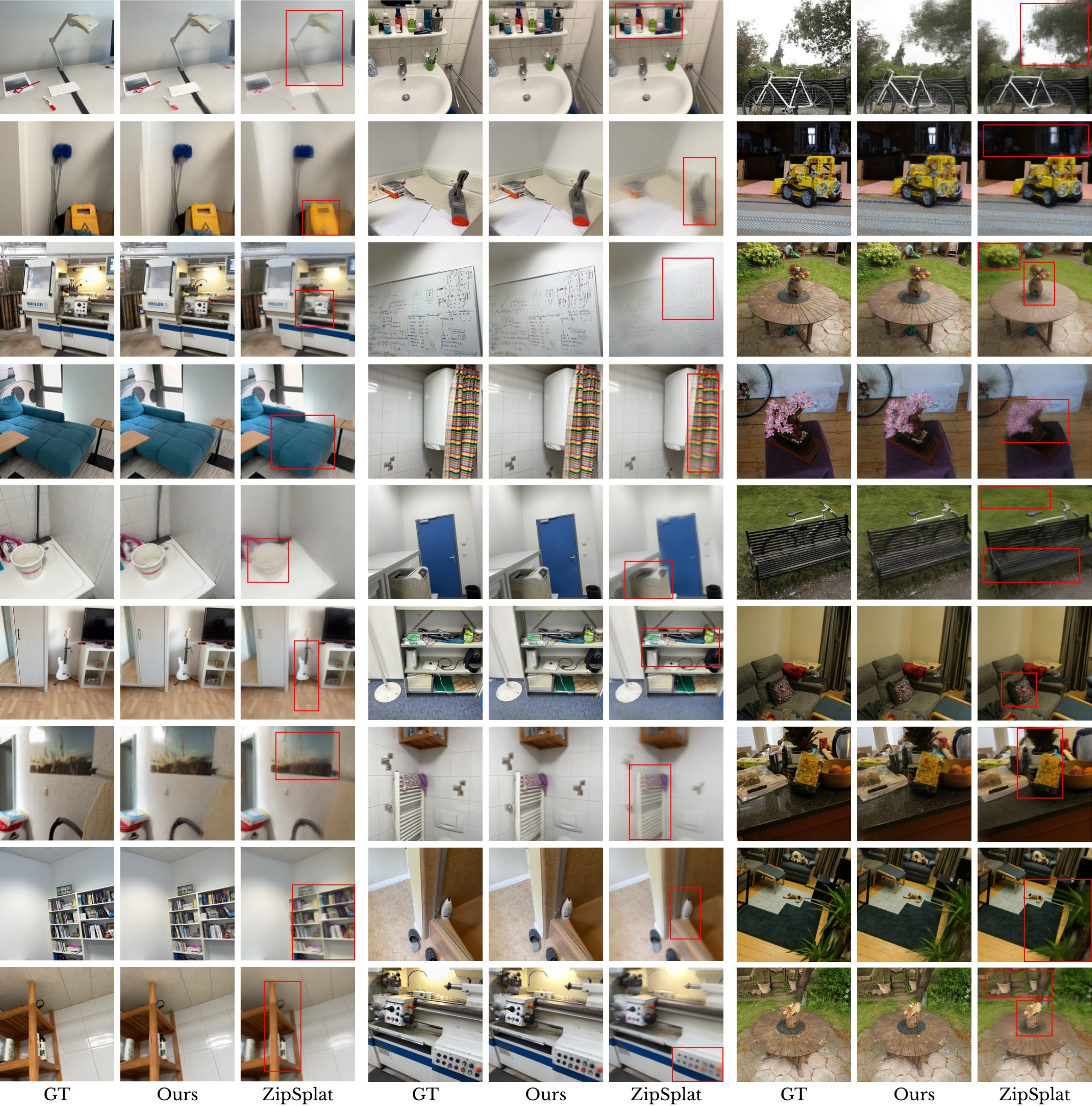}
    \caption{
        Qualitative results of cross-dataset generalization from DL3DV to ScanNet++ (the first six columns) and Mip-NeRF~360 (the last three columns). 
        Rows 1 to 3, 4 to 6, and 7 to 9 correspond to 32, 64, and 128 context views, respectively.
    }
    \label{fig:scannetpp_mip360_ablation_qualitative}
\end{figure}

\end{document}